\documentclass[a4paper]{article}
\usepackage{authblk}
\usepackage[english]{babel}
\usepackage[T1]{fontenc}
\usepackage[utf8]{inputenc}
\usepackage{textcomp}
\usepackage[a4paper,top=3cm,bottom=2cm,left=3cm,right=3cm,marginparwidth=1.75cm]{geometry}

\usepackage{xcolor}
\usepackage{amsmath}
\usepackage{graphicx}
\usepackage[colorinlistoftodos]{todonotes}
\usepackage[colorlinks=true, allcolors=blue]{hyperref}
\usepackage{booktabs}
\usepackage{siunitx}
\usepackage{graphicx}
\usepackage{hyperref}
\usepackage{listings}
\usepackage{amsmath}
\usepackage{amssymb}
\usepackage{verbatim}
\usepackage{color}
\usepackage{wrapfig}
\usepackage{array}
\usepackage{makecell}
\usepackage{tabularx}
\newcolumntype{Y}{>{\centering\arraybackslash}X}
\usepackage{subcaption}
\usepackage{caption}
\usepackage{listings}
\usepackage{xcolor}
\usepackage{colortbl}
\usepackage{hyperref}
\usepackage{authblk}
\usepackage{longtable}
\usepackage{adjustbox}
\usepackage{multirow}
\usepackage{float}
\usepackage{setspace}
\usepackage{appendix}
\usepackage{chngcntr}
\usepackage{cleveref}
\AtBeginEnvironment{appendices}{%
  \crefalias{section}{appendix}%
  \counterwithin{figure}{section}%
  \counterwithin{table}{section}%
  \counterwithin{equation}{section}%
}
\usepackage{pdfpages}
\usepackage[style=nature,url=true,doi=true,maxbibnames=3, defernumbers=true]{biblatex}
\bibliography{sample.bib}
\usepackage{pdflscape}
\newcolumntype{L}[1]{>{\raggedright\arraybackslash}p{#1}}
\newcolumntype{Y}{>{\raggedright\arraybackslash}X}

\author [1] {Talha Ansar}
\author[1]{Muhammad Mujtaba Abbas}
\author[2]{Ramit Debnath}
\author[3]{Vivek Dua}
\author[2,3,4*]{Waqar Muhammad Ashraf}

\title{\textbf{Data-driven Bi-level Optimization of Thermal Power Systems with embedded Artificial Neural Networks}}

\affil[1]{Department of Mechanical Engineering, University of Engineering and Technology Lahore, New Campus, Kala Shah Kaku, 39020, Pakistan}
\affil[2]{Collective Intelligence \& Design Group, University of Cambridge, Cambridge, UK}
\affil[3]{The Sargent Centre for Process Systems Engineering, Department of Chemical Engineering, University College London, Torrington Place, London, WC1E 7JE, UK}
\affil[4]{The Alan Turing Institute, British Library, 96 Euston Road, London, NW1 2DB, UK
}
\affil[*]{corresponding author: wma27@cam.ac.uk}

\begin{document}
\maketitle
\section*{Graphical Abstract}

\begin{figure}[htp]
    \centering
    \includegraphics[width=0.9\linewidth]{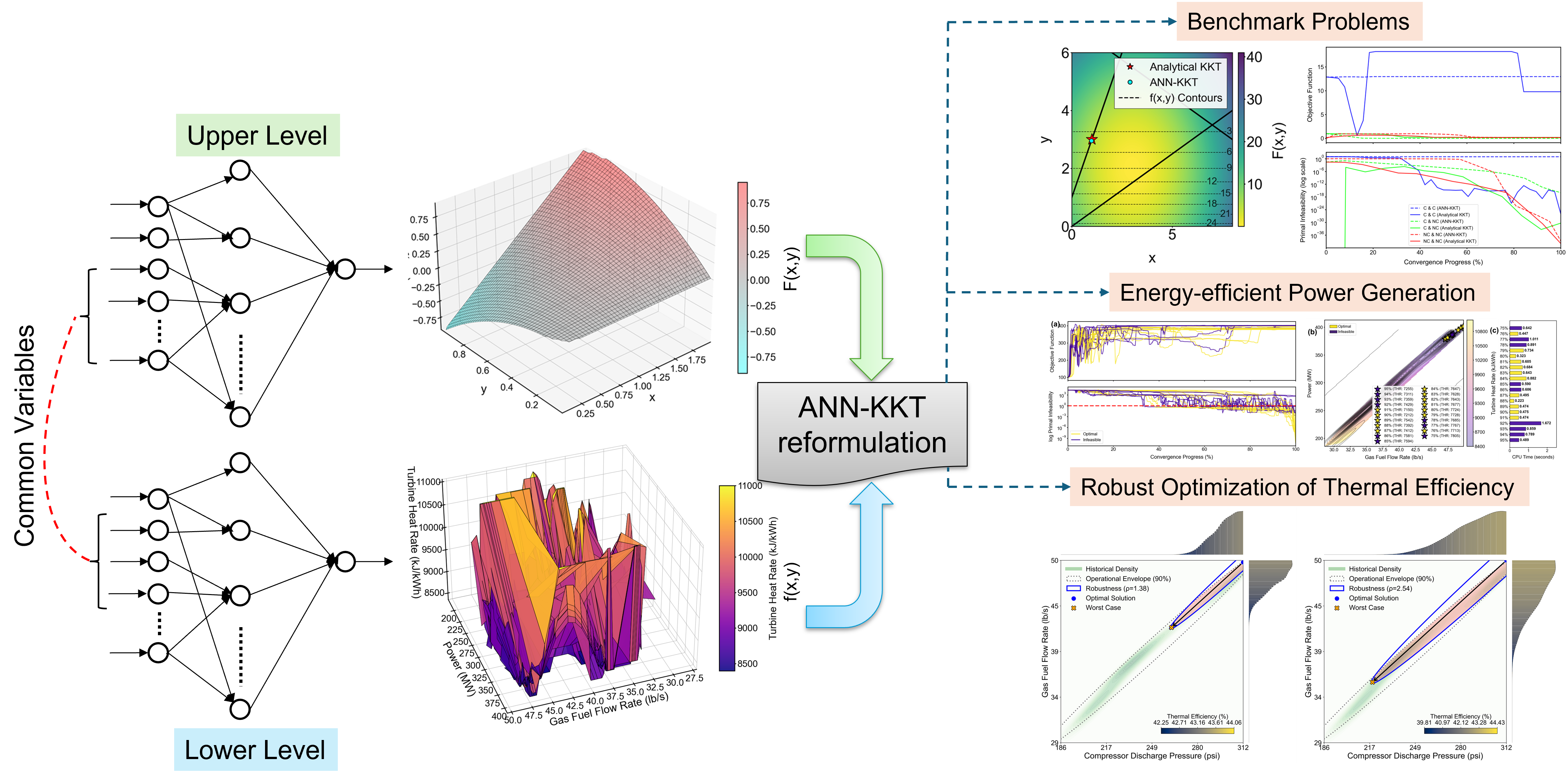}
\end{figure}

\maketitle

\begin{abstract}
Industrial thermal power systems have coupled performance variables with hierarchical order of importance, making their simultaneous optimization computationally challenging or infeasible. This barrier limits the integrated and computationally scaleable operation optimization of industrial thermal power systems. To address this issue for large-scale engineering systems, we present a fully machine learning-powered bi-level optimization framework for data-driven optimization of industrial thermal power systems. The objective functions of upper and lower levels are approximated by artificial neural network (ANN) models and the lower-level problem is analytically embedded through Karush–Kuhn–Tucker (KKT) optimality conditions. The reformulated single level optimization framework integrating ANN models and KKT constraints (ANN-KKT) is validated on benchmark problems and on real-world power generation operation of 660 MW coal power plant and 395 MW gas turbine system. The results reveal a comparable solutions obtained from the proposed ANN-KKT framework to the bi-level solutions of the benchmark problems. Marginal computational time requirement (0.22 to 0.88 s) to compute optimal solutions yields 583 MW (coal) and 402 MW (gas turbine) of power output at optimal turbine heat rate of 7337 kJ/kWh and 7542 kJ/kWh, respectively. In addition, the method expands to delineate a feasible and robust operating envelope that accounts for uncertainty in operating variables while maximizing thermal efficiency in various scenarios. These results demonstrate that ANN–KKT offers a scalable and computationally efficient route for hierarchical, data-driven optimization of industrial thermal power systems, achieving energy-efficient operations of large-scale engineering systems and contributing to industry 5.0.

\end{abstract}

\vspace{1em} 
\noindent\textbf{Keywords:} Data-driven Bi-level Optimization, Robust Optimization, Gas Power Plant, Process Flexibility, Engineering

\section{Introduction}
The operation of thermal power systems including solar-powered power generation systems, waste to power systems, fossil-powered power plants seek to optimize the performance variables related to thermodynamic, economic, safety and environment aspects. Traditionally, multi-objective optimization analysis is carried out for these competing objectives and a Pareto-optimal solution is identified which is not necessarily a single global optimum \cite{deb2001nonlinear, bejan1995thermal}. Many of the performance variables of thermal power systems have an  hierarchical order of importance. For example, plant-level economic or production output objectives from a thermal power plant are decided first (upper-level objective) and the fuel consumption in the boiler (lower-level objective) is optimized based on the decision of upper-level objective.

The natural hierarchical decision structure motivates to formulate bi-level optimization problem for thermal power systems \cite{bard2013practical}, say maximize power generation (upper level objective) given that turbine heat rate is locally or globally minimized (lower level objective). However, solving the bi-level optimization problem with competing objectives, non-convexities and couplings among the decision variables at both levels requires large computational budget and the problem can be NP-hard or the feasible solution does not exist \cite{dempe2002foundations}. Traditionally, Karush-Kuhn Tucker (KKT) optimality conditions \cite{garces2009bilevel,bard1982explicit} are used to reformulate the bi-level problem into a single level problem where the lower level problem is replaced by KKT constraints. The key limitation with KKT-based reformulation is to ensure the continuous nature of the inner problem \cite{marcotte2001bilevel}. In this regard, several alternative approaches are reported such as branch-and-bound inspired iterative methods \cite{gumucs2005global, mitsos2010global, kleniati2015generalization}, decomposition approaches \cite{saharidis2009resolution}, gradient methods \cite{sachio2022integrating, banker2025gradient}, genetic-inspired heuristics \cite{sinha2017evolutionary, hecheng2008exponential}, multi-parametric methods \cite{avraamidou2019multi, avraamidou2017multiparametric, shokry2017mixed}, and DOMINO-framework \cite{beykal2020domino}.

In the era of artificial intelligence (AI) and digitalization, the operation data of industrial thermal power systems is available, and machine learning (ML) models are routinely used to approximate the data-driven relationships between process variables and performance variables \cite{shokry2021machine}. Artificial neural network (ANN) provides a continuous function approximated from the data and is integrated into the reformulated KKT-based problem (ANN-KKT) to handle the mapping of input-output variables in high dimensions and to compute gradients for the calculation of the optimal solution \cite{molan2023using}. A key advantage of ANN-KKT reformulation is the reduced computational time required to estimate the optimal solution (local or global) \cite{medina2021reformulation} compared to the techniques mentioned earlier. This is important for the control of dynamic industrial operations where quick adjustment in the process set-points is needed for stability and safety of the engineering and thermal power processes.

The sound mathematical conception of bi-level framework integrating the ANN model is implemented for solving problems of different nature. In addition to tuning the hyperparameters of the ML models \cite{abdulkadirov2023survey}, the bi-level framework was utilized to estimate the subsidy policy making at the upper level and the design of the energy system at the lower level for the hybrid renewable energy system \cite{luo2021bi}. The lower level problem (design of the energy system) was replaced by the ANN model to compute the optimal solution. In another study, the ANN model was used as a surrogate to represent the problem at the lower-level to develop a model predictive control strategy for bioreactors \cite{de2021nonlinear}. Similarly, ANN based surrogates for lower-level problems have been reported to accelerate computational performance to compute optimal solutions \cite{bagloee2018hybrid, moreno2025solving}. Similarly, bi-level framework is implemented for solving problems by mathematical programming for system design \cite{shu2024overcoming, abraham2025multi}, planning \& scheduling \cite{leenders2023bilevel,beykal2022data,avraamidou2019b}, and parameter estimation \cite{bollas2009bilevel,mitsos2009bilevel} in chemical and energy domains.

\subsection*{Research gap}

A rich body of literature is maintained for solving bi-level nature of problems in various disciplines, the implementation of ANN-KKT framework for robust and data-driven optimization of industrial thermal power systems remains limited. Some research studies report the replacement of the lower-level problem with ANN surrogate; complete replacement of the objective functions with stand-alone ANN models in bi-level framework remains unexplored for data-driven optimization of industrial thermal power systems. This exploration as well as investigation are particularly important for scaling architectural framework of bi-level problem to solving engineering problems with hierarchical order of importance of performance variables in thermal power systems (e.g., maximizing power generation given that turbine heat rate is minimized). The computational performance of the ANN-KKT framework to compute optimal solution(s) for bi-level problem of industrial thermal power systems is worth investigating that will highlight the suitability of the framework to advance its adoption in industrial environments by effectively responding to industrial operation control dynamics. 

Another open research challenge in advancing ML in industrial operations is to establish robust operating space of the process variables for the set value of target that hedges the process uncertainty and errors in the sensor-based measurements. The current body of literature lacks in solving the robust optimization problem for the identification of operating space through ANN-KKT framework. The integration of actual data of thermal power plants for the identification of operating space under various operating scenarios is rarely reported and is a gap in the literature.    

\subsection*{Contributions of this work}

The key contributions of this work are as follows:

\begin{itemize}
    \item We present ANN-KKT framework that incorporates ANN surrogates in the objective function at both levels of the bi-level problem. The lower-level problem is replaced by KKT-conditions, and the single-level problem is reformulated,
    \item The optimization solvers may have numerical instability for satisfying the KKT constraints. We propose to reformulate them with Fischer–Burmeister function and the qualification of constraints can be verified,
    \item The ANN-KKT framework is implemented to solve benchmark problems taken from literature as well as to power generation operation of industrial thermal power plants. Power production is maximized at the upper level while turbine heat rate is minimized at the lower-level for the 666 MW coal power plant and the 395 MW gas turbine system. Mahalanobis distance-based constraints are also embedded to estimate domain-consistent optimal solutions \cite{ashraf2026domain},
    \item The ANN-KKT framework is also implemented to determine the robust operating space to hedge against process uncertainty for maximizing thermal efficiency of the gas turbine system. The framework delivers sizeable operating space for the process variables against the target efficiency floor,
    \item The computational time consumed to compute the optimal solutions for the considered problems is presented. The ANN-KKT framework requires reasonable computational time for computing the optimal solutions that can match with the industrial operation control dynamics, and
    \item The proposed ANN-KKT framework  emerges as a scaleable approach to solving bi-level problems for industrial thermal power systems and can advance AI adoption in industrial process control for smart operation management and contribute to industry 5.0
\end{itemize}

The rest of the paper is structured as follows. First, we describe the working of ANN and formulate a general-purpose bi-level problem which is reformulated with ANN-KKT framework. In the results section, we implement the ANN-KKT framework to solve benchmark problems and compare the results with the bi-level solutions. Later, ANN-KKT framework is scaled to maximize power generation from thermal power plants with the aim of minimum heat rate consumption. The framework is also scaled to robust-optimization problem to hedge the process uncertainty. Lastly, we describe the main findings of the paper in the conclusions section and highlight the limitations and future work.

\section{Methods}

In this research, we propose replacing the objective functions of bi-level problem with feed-forward ANN models. First, we define the architecture of the ANN model and optimize it to achieve good predictive accuracy for modeling tasks. Later, we formulate the bi-level optimization problem, embed the trained ANN model, and replace the lower-level problem with KKT constraints that convert the bi-level problem into a single level optimization problem (ANN-KKT). The ANN-KKT framework is implemented to solve case studies taken from the literature and power generation operation of a 660 MW coal power plant and a 395 MW capacity gas turbine system. More details about the ANN model development, reformulation of bi-level problem and specific case studies are provided in the following sub-sections.    

\subsection{Optimizing the training of ANN model}
In this research a three-layer shallow and feed-forward ANN model is considered. The shallow neural network has smaller parametric space than deep neural network, which provides reasonable computational expense for computing the gradients (generally stable than those of deep neural network) for downstream optimization tasks. Although deep neural networks have their unique application space in object detection, natural language processing, generative AI etc., many of the chemical engineering and energy systems applications can be modeled with reasonable accuracy using shallow neural networks \cite{daoutidis2024machine,dobbelaere2021machine,mowbray2022industrial} (as long as sufficient computational components are embedded in the architecture of model \cite{haykin2009neural}).

A three-layered, feed-forward ANN model consists of input, hidden, and output layers. The information from the input layer is passed to the hidden layer that has a number of computational components, formally called hidden layer neurons, to process the incoming information. The weight connections between the input-hidden layer contribute to information processing at the hidden layer through a series of operations, and the transformed signal from each hidden layer neuron is propagated forward to the output layer of the ANN model for further processing. The feed-forward information processing in the ANN model simulates a response ($\hat{y}$) which is calculated as:

\begin{equation}
\hat{y} = f_2\left(\sum W_2. f_1\left(\sum W_1.X^T + b_1\right)+ b_2\right)
\label{ANN}
\end{equation}

here, $W_1$ and $W_2$ are the weight connections between input-hidden and hidden-output layers of ANN, respectively while $b_1$, $b_2$ are the bias observations and $f_1$ and $f_2$ are activation functions applied on hidden and output layers of ANN, respectively. The model-predicted response is compared with the actual "ground truth", i.e., ($y$) and error signal is constructed. The back-propagation of error throughout the layers of ANN tunes the weight connections between the processing layers or in general model parameters ($\Theta$) \cite{rumelhart1986learning}. This is enabled through differentiating the loss function ($L$) w.r.t $\Theta$ such that $L$ is minimised and the optimal values of model parameters are computed. In this research, $L$ consists of mean-squared error and $L_1$ regularisation terms, and is written as:

\begin{equation}
L =  \frac{\sum_{i=1}^N (y_i - \hat{y}_i)^2}{N} + \lambda_1 . \Theta
\end{equation}
here, $\lambda_1$ is a hyperparameter that controls the contribution of $L_1$ regularisation to minimise the parametric values of $\Theta$. The iterative process of forward and backward propagation of information improves the pattern recognition ability of ANN model, and the model maps the data-driven relationships between the input-output variables as internal representation in the model architecture.

In order to achieve good predictive accuracy of the ANN, a number of hyperparameters are optimized that include the learning rate, the number of neurons in the hidden layer, $\lambda_1$ and $\lambda_2$ as the weight decay parameter for the optimization solver. The Bayesian optimization technique is used for the optimization of hyperparameters from their assigned search space distribution. The Tree Parzen Estimator solver, which is available in the Hyperopt library in Python \cite{bergstra2011algorithms,bergstra2013hyperopt}, is used for hyperparameter optimization. Later, the optimized hyperparameter values are embedded in the ANN architecture to tune $\Theta$. It is important to note here that Adaptive Moment Estimation (ADAM) solver \cite{kingma2014adam} is used for tuning $\Theta$ since it has stable and efficient performance for training the ANN models. We have split the data into training, testing, and validation datasets on the data split ratio of 0.7, 0.15, and 0.15, respectively, for the model development for case studies. The predictive accuracy of the models is evaluated by the coefficient of determination (R$^2$) and the root-mean squared error (RMSE) \cite{huang2025optimisation, ashraf2023machine}.

\begin{equation}
R^2 = 1 - \frac{\sum_{i=1}^N (y_i - \hat{y}_i)^2}{\sum_{i=1}^N (y_i - \bar{y_i})^2}
\end{equation}
\begin{equation}
RMSE = \sqrt\frac{\sum_{i=1}^N (y_i - \bar{y_i})^2}{N}
\end{equation}

    R$^2$ is taken as a proxy of predictive precision and varies from zero to one. However, RMSE is the mean error computed on the model-predicted responses for the given input data. Low RMSE represents the good match between the actual and model-predicted responses and, in turn, reflects the good predictive accuracy of the trained model and vise versa. Please see section \ref{sec:modelling} in the appendix to find more specific details about the training protocol for ANN model development.

\subsection{Formulation of the bi-level optimization problem}

The general formulation of bi-level optimization problem follows the leader-follower hieratical relationship where the upper level problem corresponds to the leader and the lower level problem is defined for the follower. The two problems are connected by decision variables (say $x$ for the upper level and $y$ for the lower level) incorporated into their objective functions and/or the corresponding constraints. The upper level decision variables $x$ may overlap lower level decision variables $y$ in feature space to establish the bi-level structure of the problem. The leader aims to minimize its constrained objective function $F(x, y)$ and takes a decision which is passed to the follower, which optimizes its own objective function $f(x, y)$ for the given decision of the leader and the associated constraints ($g(x, y)$, $h(x, y)$) in the problem of the lower-level.  

\begin{align}
& \underset{x \in X}{\text{minimize}}
& & F(x, y) \label{eq:upper_problem} \\
& \text{subject to}
& & H(x, y) = 0, \notag \\
& & & G(x, y) \leq 0, \notag \\
\intertext{where $y$ solves the lower-level problem:}
& y^* = \underset{y \in Y}{\text{argmin}}
& & f(x, y) \label{eq:lower_problem} \\
& \text{subject to}
& & h(x, y) = 0, \notag \\
& & & g(x, y) \leq 0, \notag \\
& & & [x_1, \dots, x_m] \in \mathbb{R}^m, \quad [y_1, \dots, y_p] \in \mathbb{R}^p. \notag
\end{align}

\subsubsection{ANN-KKT framework for bi-level problem}

A common strategy to solve bi-level problem is to reformulate it as single-level optimization problem where the lower-level problem is represented as KKT constraints for the upper level. Resultantly, the reformulated single-level optimization problem computes the local or global optimal solution for the lower-level problem depending upon its nature (convex or non-convex) and the optimization solver. In this paper, we have trained feed-forward ANN models with smooth activation functions that provide continuously differentiable expressions. These ANN model expressions replace the objective function at the two levels of bi-level problem. The objective functions are replaced with ANN models, both at upper ($F_{ANN}(x, y)$) and lower ($f_{ANN}(x, y)$) levels. The reformulated single-level with ANN-based objective functions and KKT conditions is written as:

\begin{align}
& \underset{x, y, \mu, \lambda}{\text{minimize}}
& & F_{ANN}(x, y) \notag \\
& \text{subject to}
& & H(x, y) = 0, \\
& & & G(x, y) \leq 0, \label{eq:upper_const} \\
\intertext{{Stationarity (Lagrangian Gradient):}}
& & & \nabla_y f_{ANN}(x, y) + \mu^T \nabla_y g(x, y) + \lambda^T \nabla_y h(x, y) = 0, \label{eq:stationarity} \\
\intertext{{Primal Feasibility:}}
& & & g(x, y) \leq 0, \label{eq:primal_ineq} \\
& & & h(x, y) = 0, \label{eq:primal_eq} \\
\intertext{{Dual Feasibility:}}
& & & \mu \geq 0, \label{eq:dual_feas} \\
\intertext{{Complementary Slackness:}}
& & & \mu . g(x, y) = 0, \label{comp_1}\\
& & & [x_1, \dots, x_m] \in \mathbb{R}^m, \quad [y_1, \dots, y_p] \in \mathbb{R}^p. \label{compl}
\end{align}

The stationarity condition computes the Lagrangian function gradient and requires the lower-level problem to be strictly continuous. The qualification of KKT-based constraints written for lower-level problem only ensures local optimality of the solution computed for the bi-level problem. Equation \ref{eq:stationarity} is the specific representation of the stationarity condition that is written for the ANN model-based bi-level problem of the lower level. The KKT conditions (\ref{eq:stationarity} - \ref{compl}) are embedded with the upper level problem (\ref{eq:upper_const}) for solving ANN-KKT based reformulated bi-level problem.

\subsubsection{Reformulating complementarity conditions with Fischer-Burmeister Function for ANN-KKT framework}

The optimization solvers may face numerical instability to compute optimal solution for non-convex optimization problems involving complementarity constraints. Fischer-Burmeister (FB) function as nonlinear complementary problem (NCP) is used for reformulating complementarity conditions only when the complementarity constraint violations are observed for ANN-KKT based reformulation of bi-level problem. The FB function is mathematically written as:

\begin{equation}
\Phi_{FB}(a,b) = a + b - \sqrt{a^2 + b^2} , \quad \Phi: R^2 \rightarrow R \label{eq:fb_bound} 
\end{equation}

$\Phi_{FB}$ as NCP function has the property that $\Phi_{FB}(a,b)=0$ if and only if $a \geq 0, b \geq 0,$ and $ab=0$. Moreover, the function $\Phi_{FB}$ has another property that $\Phi_{FB}(a,b)=0$ and $a,b \geq0$ is equivalent to $0\leq a \perp b \leq 0$ \cite{leyffer2006complementarity}. This property is used to formulate a constraint based on functions $\Phi_{FB}$ to replace the complementarity slackness of the KKT conditions when solvers face numerical instability to arrive at an optimal solution for the optimization problems considered for case studies. However, $\Phi_{FB}$ at $a=b=0$ is not differentiable; its perturbed variant ($\Phi_{FB}^*(a,b,\epsilon)$) satisfies $\Phi_{FB}(a,b,\epsilon)=0$ if and only if $a \geq 0, b \geq 0$, $ab=\epsilon/2$ for $\epsilon > 0$ \cite{lv2008neural}. $\Phi_{FB}^*(a,b,\epsilon)$ is smooth with respect to $a$ and $b$ when $\epsilon > 0$ and is written as: 

\begin{equation}
\Phi_{FB}^*(a,b,\epsilon) = \sqrt{a^2 + b^2 + \epsilon } - a - b , \quad \Phi: R^3 \rightarrow R \label{eq:fbp_bound} 
\end{equation}

Initially, KKT-transformed bi-level problems involving complementarity constraints are solved without reformulating the complementarity constraint through $\Phi_{FB}^*(a,b,\epsilon)$. In case of constraints violations for lower-level complementarity constraints (Equation \ref{comp_1}), they are reformulated by $\Phi_{FB}^*(a,b,\epsilon)$. In this case, primal feasibility and dual feasibility constraints are not embedded separately in ANN-KKT problem as complementarity constraint reformulated by $\Phi_{FB}^*(a,b,\epsilon)$ inherently ensures compliance of primal and dual feasibility constraints. The Linear Independence Constraint Qualification (LICQ) is evaluated for the reformulated problem at solution by verifying the full rank of the Jacobian matrix of active constraints. If LICQ holds, it indicates that the gradients of the active constraints are linearly independent, which guarantees the uniqueness of the associated Lagrange multipliers and confirms that the solution satisfies the necessary KKT conditions for optimality \cite{NocedalWright2006}.

 The ANN-KKT problem with $\Phi_{FB}^*(a,b,\epsilon)$ based reformulation of complementarity constraint is given in the following:

\begin{align}
& \underset{x, y, \mu, \lambda}{\text{minimize}}
& & F_{ANN}(x, y) \notag \\
& \text{subject to}
& & H(x, y) = 0, \\
& & & G(x, y) \leq 0, \label{eq:upper_const_fb} \\
\intertext{{Stationarity (Lagrangian Gradient):}}
& & & \nabla_y f_{ANN}(x, y) + \mu^T \nabla_y g(x, y) + \lambda^T \nabla_y h(x, y) = 0, \label{eq:stationarity_fb} \\
\intertext{{Complementary Slackness:}}
& & & \Phi_{FB}^*(\mu_g,g(x, y),\epsilon) = \sqrt{\mu_g^2 + (g(x, y))^2 + \epsilon } - \mu_g - g(x, y) = 0, \\
& & & \Phi_{FB}^*(\mu_y,y,\epsilon) = \sqrt{\mu_y^2 + y^2 + \epsilon}  - \mu_y - y = 0, \label{box_comp_fb}\\
& & & [x_1, \dots, x_m] \in \mathbb{R}^m \label{compl_fb}
\end{align}

As described above, $\mu_g$ and $g(x,y)$ complement each other and by enforcing $\Phi_{FB}^*(\mu_g,g(x, y),\epsilon) = 0$, it is implied that $\mu_g, g(x,y) \geq 0$ and $\mu_g \cdot g(x,y) = \epsilon/2$ for $\epsilon >0$. Consequently, reformulation is written such that $g(x,y)$ returns $\mathbb{R}^+$ which is consistent with the nature of the constraint and ensures equivalence with $\Phi_{FB}^*(\mu_g,g(x, y), \epsilon)$. Similarly, the reformulated complementarity constraint on the bounds of the lower-level decision variables ($y$) can be expressed by Equation \ref{box_comp_fb}, where $y$ is scaled to $\mathbb{R}^+$.

In this paper, the bi-level analysis is carried out on the following operating system specification: Intel® CoreTM i7-8850H processor (6 cores, 12 threads, base frequency 2.6 GHz, turbo up to 4.3 GHz), 32 GB DDR4 memory (2 × 16 GB, 2667 MHz), a 512 GB NVMe Gen 4 solid-state drive, and an NVIDIA Quadro P1000 GPU with 4 GB GDDR5 VRAM. The operating system environment was Windows 10 Pro (64-bit). However, the versions of different libraries used in the development of ML models and optimization framework are as follows: pandas 2.2.3, torch 2.6.0+cpu, numpy 1.26.4, scikit-learn 1.6.1, hyperopt 0.2.7, pyomo 6.9.0 and python 3.9.

\section{Results}

We implement the ANN-KKT framework for solving case studies taken from the literature.  Moreover, we also analyze the maximum power generation from thermal power plants (660 MW coal power plant and 395 MW gas turbine system) with the ANN-KKT framework. The ability of ANN-KKT framework for robust optimization problem written for maximizing the thermal efficiency of gas turbine system under different scenarios is also investigated. The details of the results for the case studies are provided in the following.

\subsection{Benchmark Examples}
Three benchmark examples are adopted from the literature to capture distinct objective function landscapes across the upper and lower levels of the problem, specifically: convex \& convex, convex \& non-convex, and non-convex \& non-convex cases. The mathematical formulations for these benchmarks are detailed in Section \ref{examp_1} -- Section \ref{exam_3}, with results discussed in Section \ref{res_exam}. To develop the ANN approximation of these problems, synthetic datasets are generated for each problem instance. The input space is explored using a uniform sampling strategy $U[\min, \max]$ to ensure adequate coverage of the decision variables. A total of 10,000 samples are generated for each problem and subsequently divided into a training set of 7,000 samples (70$\%$), a validation set of 1,500 samples (15$\%$) and an independent testing set of 1,500 samples (15$\%$) to rigorously evaluate the generalization performance of the models. The predictive performance of trained ANN models is provided in Appendix \ref{sec:bench_extended} (Table \ref{tab:ann_performance}). The benchmark problems as well as their ANN-KKT reformulations are solved by BARON and IPOPT solvers, which are made available from GAMS.

\subsubsection{Convex \& Convex (C \& C)} \label{examp_1}

The SC2 problem \cite{mitsos2006testset} represents a standard convex-convex bi-level optimization problem. The upper-level objective seeks to minimize the distance to the point $(3,2)$, while the lower-level minimizes a quadratic function of $y$ subject to linear coupling constraints.

\begin{align}
& \underset{x \in X}{\text{minimize}}
& & F(x, y) = (x-3)^2 + (y-2)^2 \label{eq:upper_problem} \\
& \text{subject to}
& & 0 \leq x \leq 8, \notag \\
& \underset{y \in Y}{\text{argmin}}
& & f(x, y) = (y-5)^2 \label{eq:lower_problem} \\
& \text{subject to}
& & g_1(x, y) = -2x + y - 1 \leq 0, \notag \\
& & & g_2(x, y) = x - 2y \leq 0, \notag \\
& & & g_3(x, y) = x + 2y - 14 \leq 0, \notag \\
& & & 0 \leq y \leq 6, \notag \\
& & & x \in \mathbb{R}, \quad y \in \mathbb{R}. \notag
\end{align}

The problem is first solved using KKT conditions (Bi-level-KKT). Later, the problem is reformulated with the ANN-KKT framework and is written in the following:

Objective functions:
\begin{equation}
\begin{gathered}
    \min_{x, y, \boldsymbol{\lambda}, \boldsymbol{\mu}} \quad \hat{F}_{\text{ANN}}(x,y) \\
    \text{here }\qquad \hat{F}_{\text{ANN}}(x,y) \approx (x-3)^2 + (y-2)^2 \\
     \qquad \hat{f}_{\text{ANN}}(x,y) \approx (y-5)^2
\end{gathered}
\label{eq:sc2_mpec_obj}
\end{equation}

Upper-Level Constraints:
\begin{equation}
x \in [0, 8] \label{eq:sc2_upper_bounds}
\end{equation}

Primal Feasibility:
\begin{align}
-2x + y - 1 &\leq 0 \label{eq:sc2_primal_1}\\
x - 2y &\leq 0 \label{eq:sc2_primal_2}\\
x + 2y - 14 &\leq 0 \label{eq:sc2_primal_3}\\
y &\in [0, 6] \label{eq:sc2_primal_bounds}
\end{align}

Stationarity Condition:
\begin{equation}
\nabla_{y} \hat{f}_{\text{ANN}}(y) + \lambda_1 - 2\lambda_2 + 2\lambda_3 - \mu^L + \mu^U = 0 \label{eq:sc2_stationarity}
\end{equation}

where the gradient of the lower-level neural network surrogate is derived via symbolic differentiation \cite{yeung2010sensitivity}: 
\begin{equation}
\nabla_{y} \hat{f}_{\text{ANN}}(y) = \sum_{j=1}^{H} W_{j}^{(2)} \cdot \sigma'(W_{j,y}^{(1)} \cdot y + b_{j}^{(1)}) \cdot W_{j,y}^{(1)}
\end{equation}

$W_{j,y}^{(1)}$ is the weight connection matrix from input to hidden layer having $H$ number of neurons, while $W_{j}^{(2)}$ is the weight connection matrix from hidden to output layer of ANN. $\sigma'$ is the derivative of activation function applied on the hidden layer of ANN. $\mu^L$ and $\mu^U$ are the corresponding multipliers of box constraint ([0,6]) which are split into two inequalities.

Dual Feasibility:
\begin{align}
\lambda_1, \lambda_2, \lambda_3 &\geq 0 \label{eq:sc2_dual_lambda}\\
\mu^L, \mu^U &\geq 0 \label{eq:sc2_dual_mu}
\end{align}

Complementarity Conditions:
\begin{align}
\lambda_1 \cdot (-2x + y - 1) &= 0 \label{eq:sc2_comp_1}\\
\lambda_2 \cdot (x - 2y) &= 0 \label{eq:sc2_comp_2}\\
\lambda_3 \cdot (x + 2y - 14) &= 0 \label{eq:sc2_comp_3}\\
\mu^L \cdot -y &= 0 \label{eq:sc2_comp_lower}\\
\mu^U \cdot (y - 6) &= 0 \label{eq:sc2_comp_upper}
\end{align}

The details about the computation of solution as well as the CPU time requirement are provided in section \ref{res_exam}.

\subsubsection{Convex \& Non-convex (C \& NC)}

This problem is taken from \cite{mitsos2008global}, and features a convex upper-level objective while a non-convex lower-level objective function. The lower-level non-convexity arises from the indefinite quadratic term, $-y^2 + xy$.

\begin{align}
\intertext{{upper-level problem:}}
& \min_{x}
& & (x-1)^2 + y^2 \notag \\
& \text{subject to}
& & x \in [0, 2] \label{eq:upper_const1} \\
\intertext{{lower-level problem:}}
& \min_{y}
& & -y^2 + xy \notag \\
& \text{subject to}
& & y \in [0, 1], \label{eq:lower_const1}
\end{align}

The ANN-KKT reformulation of the problem is mentioned in section \ref{ANN_KKT_B1}. The details about the computation of solution as well as the CPU time requirement are provided in section \ref{res_exam}.

\subsubsection{Non-convex \& Non-convex (NC \& NC)} \label{exam_3}

The "Pathological Branching" problem (mb\_1\_1\_16) \cite{mitsos2006testset} is a challenging test case where both upper and lower levels are non-convex. It includes a difficult constraint, ($y^2(x-0.5) \leq 0$), in the lower level that tests the ability of the solver to handle singularities and non-standard geometries. The problem is defined in the following:

\begin{align}
\intertext{{upper-level problem:}}
& \min_{x,y}
& & x^2 \notag \\
& \text{subject to}
& & 1 + x - 9x^2 - y \leq 0, \label{eq:upper_const1} \\
& & & -1 \leq x \leq 1. \notag \\
\intertext{{lower-level problem:}}
& \min_{y}
& & y \notag \\
& \text{subject to}
& & y^2(x - 0.5) \leq 0, \label{eq:lower_const1} \\
& & & -1 \leq y \leq 1. \notag
\end{align}.

The ANN-KKT reformulation of the problem is mentioned in section \ref{ANN_KKT_B2}.

\subsubsection{Results of Benchmark Problems} \label{res_exam}

\begin{figure}[H]
    \centering
    \includegraphics[width=0.9\linewidth]{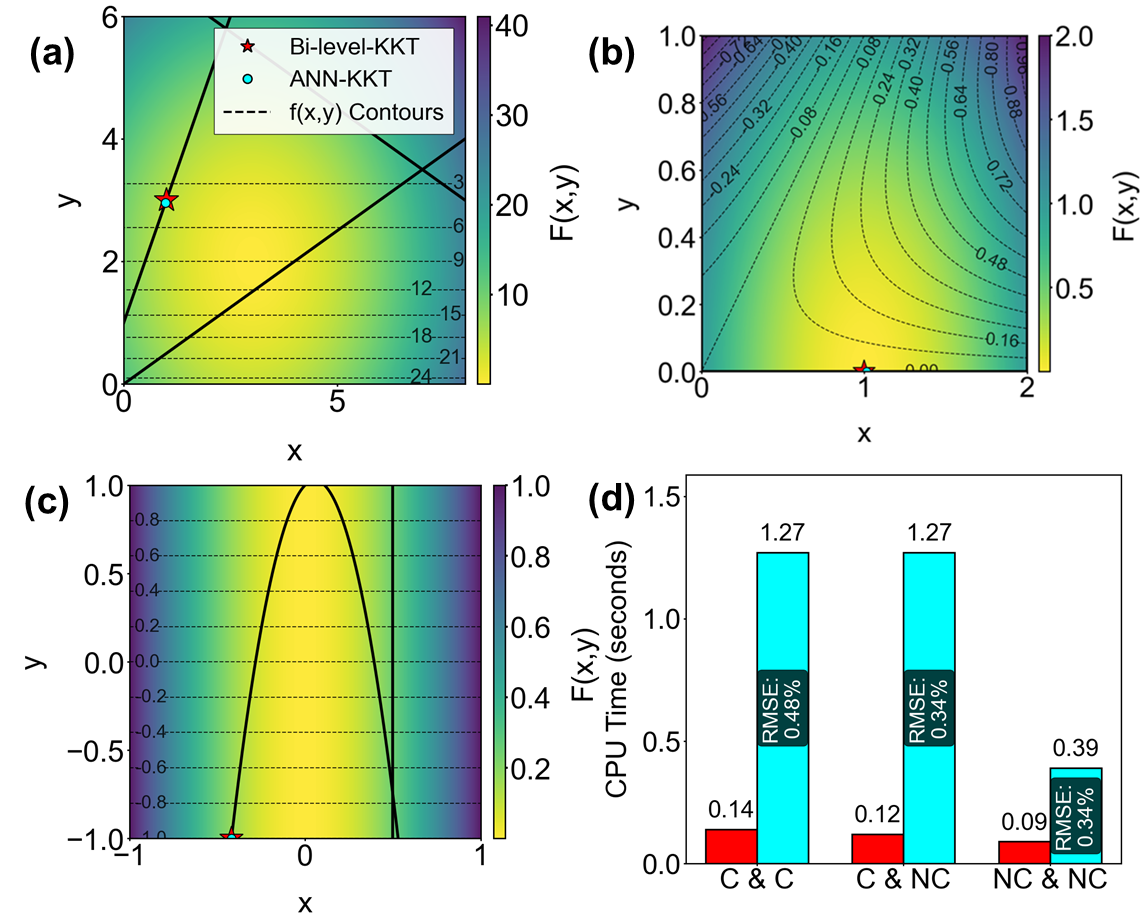}
    \caption{Visualizations of benchmark problems optimization landscape for problems of nature (a) C \& C, (b) C \& NC, and (c) NC \& NC. The background gradient represents the upper-level objective function value, while the dashed lines denote the lower-level objective. Solid black lines define the feasible solutions. (d) CPU time requirement for solving the benchmark problems - red: Bi-level-KKT and blue: ANN-KKT. The approximation error of ANN-KKT framework is also mentioned.}
    \label{fig:SC_2}
\end{figure}

Bi-level-KKT and ANN-KKT reformulations of the benchmark problems (C $\&$ C, C $\&$ NC, and NC $\&$ NC) are solved using the BARON solver. Figure \ref{fig:SC_2} and Table \ref{tab:merged_results} present the comparison of solutions obtained via the ANN-KKT framework and Bi-level-KKT against the bi-level solutions. For the three benchmark problems, the ANN-KKT approach successfully approximates the solutions with negligible relative absolute error in comparison to bi-level solutions (refer to Figure \ref{fig:SC_2}(d)). Although the ANN-KKT formulation incurs higher computational costs (CPU time ranges from 0.39 s to 1.27 s) compared to the Bi-level-KKT (0.09 s to 0.14 s) due to the added complexity of computing neural network gradients, it demonstrates reliability as a surrogate for computing the optimal solution for complex bi-level topologies. 

\begin{table}[H]
\centering
\caption{Comparison of solutions and CPU times for solving different bi-level problems (Convex \& Convex, Convex \& Non-Convex, Non-Convex \& Non-Convex).}
\label{tab:merged_results}
\begin{tabular}{llcccc}
\toprule
\textbf{Problem Type} & \textbf{Method} & \textbf{x} & \textbf{y} & \textbf{Objective (F)} & \textbf{Time (s)} \\
\midrule
\multirow{3}{*}{C \& C} 
 & {Bi-level} \cite{mitsos2006testset} & 1.0 & 3.0 & 5.0 & - \\
 & Bi-level-KKT & 1.0 & 3.0 & 5.0 & 0.14 \\
 & ANN-KKT & 0.981 & 2.962 & 4.976 & 1.27 \\
\cmidrule{1-6}

\multirow{3}{*}{C \& NC} 
 & {Bi-level} \cite{mitsos2008global} & 1.0 & 0.0 & 0.0 & - \\
 & Bi-level-KKT & 1.0 & 0.0 & 0.0 & 0.12 \\
 & ANN-KKT & 1.014 & 0.0 & 0.0034 & 1.27 \\
\cmidrule{1-6}

\multirow{3}{*}{NC \& NC} 
 & {Bi-level} \cite{mitsos2006testset} & -0.4191 & -1.0 & 0.1756 & - \\
 & Bi-level-KKT & -0.4191 & -1.0 & 0.1756 & 0.09 \\
 & ANN-KKT & -0.4191 & -1.0 & 0.1750 & 0.39 \\
\bottomrule
\end{tabular}
\end{table}

Furthermore, the three benchmark problems are also solved using IPOPT to investigate the ability of the solver for computing the optimal solutions. The solution convergence and primal infeasibility profiles of the solver for ANN-KKT and Bi-level-KKT reformulations are provided in Appendix \ref{bench-prob} (Figure \ref{fig:bench_convergence} and Table \ref{tab:merged_results_all}), demonstrate that IPOPT converges to the optimal solutions for the NC \& NC problem with ANN-KKT and Bi-level-KKT reformulations. However, IPOPT did not converge to optimal solution for C \& C problem reformulations and Bi-level-KKT reformulation for C \& NC problems. The solution infeasibility for C \& C problem arisen from IPOPT can be reasoned by the introduction of neural network surrogates in the optimization problem which can create local minima and non-convex regions in the feasible space, even for problems that are originally convex. For C \& NC (ANN-KKT reformulation) and NC \& NC problems, IPOPT converges to optimal solutions with iteration counts ranging from 13 to 22 and final objective values closely matching those obtained with the bi-level solutions (detailed results in Appendix \ref{bench-prob} (Table \ref{tab:merged_results_all}). The results demonstrate the ability of ANN-KKT reformulation and IPOPT solver to compute optimal solution for non-convex nature of bi-level problem defined at the upper and lower levels. This aspect is further investigated for solving bi-level problems defined for the operation optimization of thermal power plants since ANN models approximated on the noisy and controlled operation data of industrial thermal power systems introduce non-convexity in the function space. However, the successful implementation of ANN-KKT framework for benchmark problems paves the way for its scalability to data-driven optimization of thermal power systems. More details about the problem formulation and obtained solution (feasible or infeasible) are provided in the following sections.

\subsection{Bi-level optimization of thermal power plants}
We further expand the implementation of ANN-KKT approach for maximizing the power generation from industrial thermal power plants such that turbine heat rate has been minimized. The historical operation data of the power generation operation of thermal power plants (660 MW coal power plant and 395 MW gas turbine system) is used to develop ANN based surrogate models which are embedded in the ANN-KKT framework. We purposely used IPOPT solver for solving reformulated ANN-KKT problem for thermal power systems for the following reasons: (i) the computation of local optimal solution in a reasonable time frame (within a few seconds) is sufficient to maintain the process dynamics, (ii) IPOPT solver is available under open-access and has established strengths and mathematical rigor to compute solution for nonlinear optimization problems \cite{wachter2006implementation}. Moreover the ANN-KKT framework can be scaled to commercial solvers for industrial problems, and (iii) IPOPT has comparable performance with many of state-of-the-art solvers \cite{rojas2015benchmarking,lavezzi2022nonlinear}.

\subsubsection{660 MW Thermal Power Plant}

The operating (decision) variables are selected from the operating space of a 660 MW coal-fired power plant for data-driven modeling of key performance indicators, namely power generation (Power, MW) and turbine heat rate (THR, kJ/kWh). In \cite{ashraf2024driving}, eight operating variables are identified for power generation operation: coal flow rate (CFR, t/h), air flow rate (AFR, t/h), main steam pressure (MSP, MPa), main steam temperature (MST, $^{\circ}$C), main steam flow rate (MSF, t/h), reheat steam temperature (RHST, $^{\circ}$C), feedwater temperature (FWT, $^{\circ}$C), and condenser vacuum (CV, kPa). All input variables are normalized to the range $[0,1]$ using min--max scaling to ensure numerical stability and efficient convergence during gradient-based optimization.

The THR model is constructed using all eight operating variables, whereas the Power model is defined on a subset of variables that excludes RHST and CV. The overall problem is formulated as a bi-level optimization framework in which the upper level seeks to maximize power generation, while the lower level minimizes THR subject to operational constraints. The dataset is taken from \cite{ashraf2024driving}, and 1,279 samples are randomly divided into a training set (70$\%$) for model parameter learning and a validation set (30$\%$) for hyperparameter tuning and early stopping to mitigate over-fitting. Whereas, an independent testing set of 222 samples is reserved for final performance assessment of trained models.  

Figure \ref{fig:data_surface} presents the operational landscape of THR as a function of CFR and Power. The surface exhibits pronounced non-convexity with multiple peaks and troughs across the operating region, confirming the highly nonlinear relationship between THR and operating variables (refer to appendix \ref{vis_conv} to compare the surface plot with those of benchmark problems). The ANN surrogates trained on the operational data capture the complex functional relationship, making bi-level problem of NC \& NC nature for coal power plant. While the KKT-reformulated single-level optimization framework provides necessary conditions for optimality, it does not guarantee global optimality for non-convex lower level problem. However, any KKT point satisfying all necessary constraints represents a feasible operating point for the operation of power plant. In this sense, locally optimal solutions satisfying the embedded constraints in the optimization problem are operationally feasible for dynamic operation of power plant, as the objective is to identify stable, efficient, and physically realizable operating points rather than the theoretical global optimum.

\begin{figure}[H]
    \centering
    \includegraphics[width=0.5\linewidth]{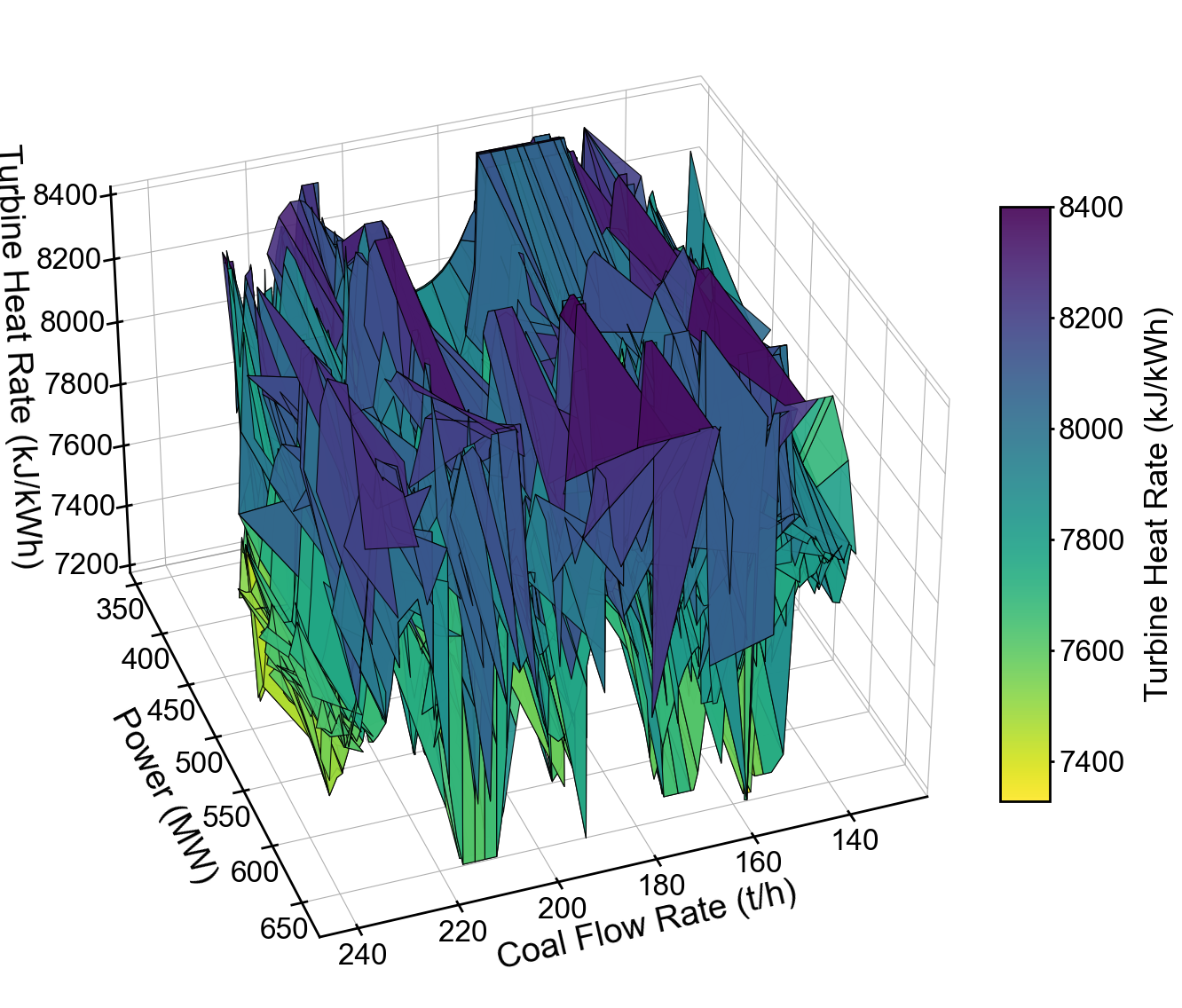}
    \caption{3D visualization of the turbine heat rate response surface as a function of coal flow rate and power output. The surface exhibits significant non-convexity with multiple local optima, characterized by irregular ridges and valleys across the operating domain. The complex topography exhibits the nonlinear interdependencies between the variables of thermal power plant. Note that this visualization represents the raw data surface rather than a fitted model profile and it serves a purely visual purpose to illustrate the data's complexity and is not an accurate depiction of the isolated relationship between these variables.}
    \label{fig:data_surface}
\end{figure}

\paragraph{Bi-level problem formulation:}

The bi-level optimization problem for thermal power plant is formulated as:

\begin{align}
\intertext{{Upper-level problem:}}
& \max_{x}
& & \hat{F}_{\text{ANN\_Power}}(x) \\
& \text{subject to}
& & x \in [0,1]^6, \quad x \in y,\notag \\
\intertext{{Lower-level problem:}}
& \min_{y}
& & \hat{f}_{\text{ANN\_THR}}(y) \notag \\
& \text{subject to}
& & (y - \mu)^T \Sigma^{-1} (y - \mu) \leq \tau^2, \label{eq:tpp_lower_maha} \\
& & & y \in [0,1]^8. \notag
\end{align}

Here, $\hat{F}_{\text{ANN\_Power}}(x)$ represents the ANN surrogate model for power output (MW), while $\hat{f}_{\text{ANN\_THR}}(y)$ represents the ANN surrogate model for THR (kJ/kWh). The Mahalanobis distance constraint ($(y - \mu)^T \Sigma^{-1} (y - \mu) \leq \tau^2$) ensures that the optimal solution remains within the valid operational envelope defined by historical training data, where $\mu$ is the mean vector, $\Sigma^{-1}$ is the inverse covariance matrix, and $\tau$ corresponds to the percentile of the Mahalanobis distance. This constraint prevents computing the solution into physically unrealistic operating regimes and it complies with the operational safety margins. The reformulated bi-level problem with KKT conditions is provided in Appendix \ref{kkt_coal} (section \ref{kkt_coal_plant}). The predictive accuracy of the trained ANN models to model Power and THR is mentioned in Appendix \ref{kkt_coal} (Table \ref{tab:coal_performance}).

Figure \ref{fig:bilevel_solutions}(a) illustrates the convergence behavior of the IPOPT solver for the bi-level optimization problem across different $\tau$ levels. Figure \ref{fig:bilevel_solutions}(a)-top panel shows the evolution of the objective function (power output) as a function of convergence progress, normalized from 0\% to 100\%. Feasible solutions (shown with yellow-green color) converge smoothly to stable objective values between 380-590 MW, with most trajectories exhibiting characteristic plateaus followed by final descent to the optimal value. Infeasible solutions (shown with dark purple color) display erratic convergence patterns with significant oscillations and premature stagnation at suboptimal objective values. Figure \ref{fig:bilevel_solutions}(a)-bottom panel presents the primal infeasibility (constraint violation) on a logarithmic scale, demonstrating the solver's ability to reduce constraint violations over the optimization trajectory. Feasible solutions successfully achieve primal infeasibility below $10^{-6}$ by the end of convergence, while infeasible solutions plateau at approximately $10^0$ to $10^1$, indicating persistent constraint violations. The red dashed line at $10^0$ serves as a visual reference for constraint violation and primal infeasibility trajectories below the red dashed line correspond to feasible solution. The distinct separation between feasible and infeasible trajectories confirms that the Mahalanobis distance tolerance directly impacts the problem's feasibility landscape, with certain percentile values creating constraint configurations that prevent convergence to optimal solutions.

\begin{figure}[H]
    \centering
    \includegraphics[width=0.8\linewidth]{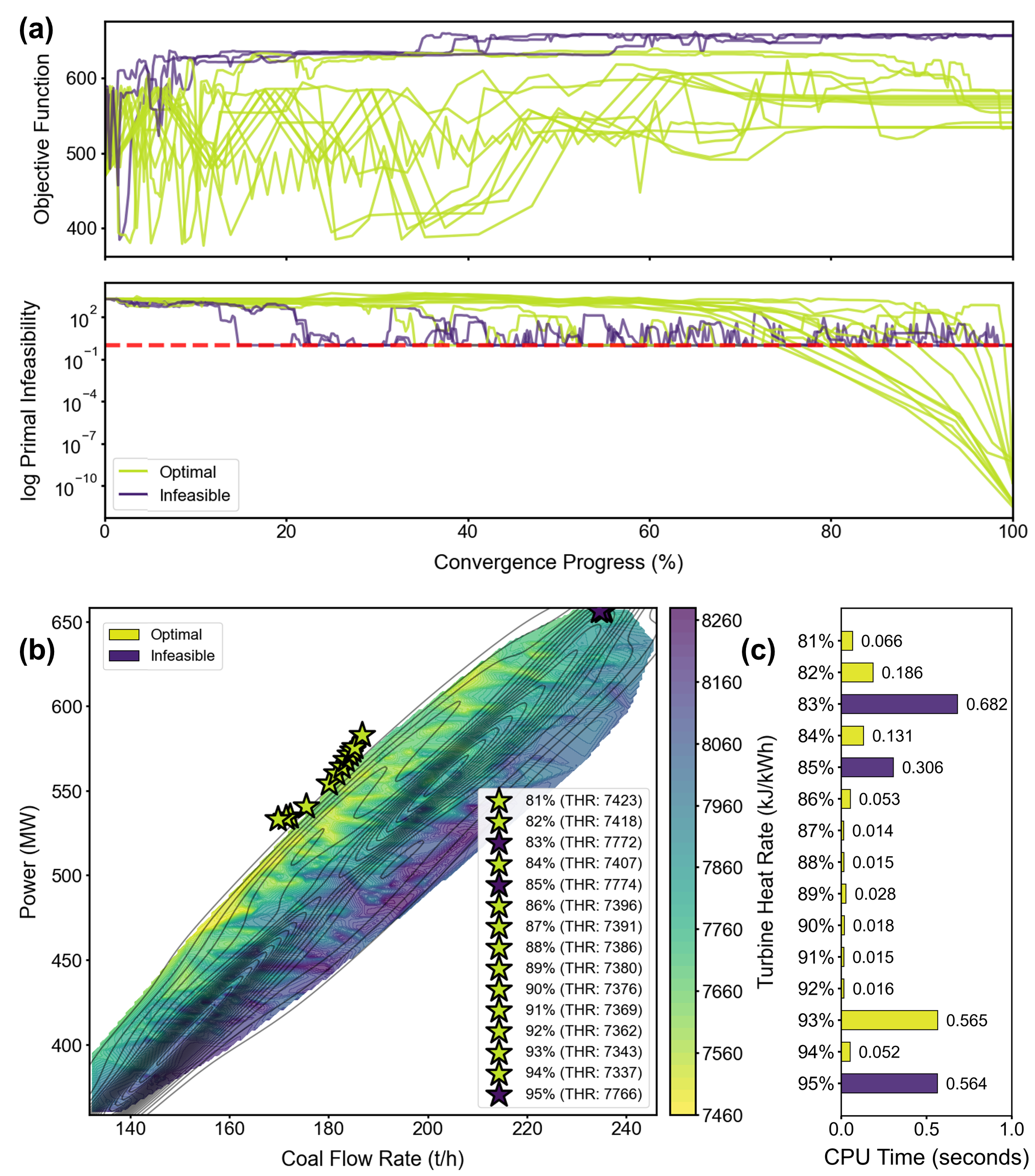}
    \caption{Optimal operating points from bi-level optimization across Mahalanobis distance tolerances. (a) IPOPT convergence profiles for objective function and primal infeasibility constraint for bi-level optimization across tolerance levels. Primal infeasibility trajectories on logarithmic scale, with feasible solutions achieving constraint satisfaction below $10^{-6}$ and infeasible solutions stagnating above $10^0$ (red dashed reference line). (b) The surface shows turbine heat rate as a function of coal flow rate and power output. Star markers indicate optimal solutions for tolerance levels from 81\% to 95\%, with yellow stars representing feasible solutions (lower THR) and purple stars indicating infeasible solutions at constraint boundaries (higher THR). Black contour lines show historical operational density. (c) CPU time consumed to compute solutions for different values of $\tau$ for Mahalanobis constraint.}
    \label{fig:bilevel_solutions}
\end{figure}

Figure \ref{fig:bilevel_solutions}(b) presents the solutions obtained from the ANN-KKT reformulated bi-level optimization across different Mahalanobis distance tolerance percentiles (81\% to 95\%). The surface represents the THR landscape as a function of coal flow rate and power output, interpolated from historical operational data. Each star marker indicates a solution (feasible, infeasible) corresponding to a specific $\tau$ level, with the color gradient reflecting the THR values. Solutions at lower tolerance percentiles (81\%-82\%) cluster in regions of lower THR (7419 kJ/kWh to 7423 kJ/kWh) while remaining optimal for maximizing power generation, reaching 551 MW to 554 MW. As the tolerance for Mahalanobis distance-based constraint increases (moving toward 95\%), the feasible operating envelope expands, allowing the solver to compute solutions from relatively wide search space. Notably, solutions at 83\%, 85\%, and 95\% percentiles exhibit higher THR values and are infeasible since constraints violations are observed. More details about the optimization results with respect to different $\tau$ levels are mentioned in appendix \ref{kkt_coal_plant} (Table \ref{tab:coal_solutions}).

Out of the feasible optimal solutions, the best solution is obtained on the $\tau$ level of 94\% that maximizes power to 583 MW with THR of 7337 kJ/kWh. The CPU time consumed to compute the optimal solution is 0.334 s and is also depicted on Figure \ref{fig:bilevel_solutions}(c). Apart from a few infeasible solutions, the CPU time remains less than one second, demonstrating the excellent computational performance of ANN-KKT framework to compute optimal solution for non-convex problem of power generation from coal power plant. The black contour lines represent the kernel density estimate of historical operating conditions, showing that most optimal solutions obtained on different $\tau$ levels (yellow colored) align well with frequently-visited operational regimes in the plant's operation. This highlights the implementability of the computed solutions for bi-level problem on the existing control layer of the power plant for the operation optimization for power generation.

\subsubsection{395 MW Gas Turbine System}
Two performance variables of gas turbine system, i.e., power generation (Power - MW) and turbine heat rate (THR - kJ/kWh) are modelled on historical operation data associated with operating variables as described in \cite{ashraf2024driving}. Following nine key operating variables are identified: compressor discharge pressure (CDP - Psi), gas fuel flow rate (GFFR - lb/s), ambient temperature (AT - $^{\circ}$F), ambient pressure (AP - hPa), ambient humidity (AH - \%), compressor discharge temperature (CDT - $^{\circ}$F), flue gas temperature at HRSG inlet (FGTI - $^{\circ}$C), fuel gas temperature (FGT - $^{\circ}$F), and performance heater gas outlet temperature (PHGOT - $^{\circ}$F). All variables are normalized to the range [0,1] through min-max scaling. The THR model is built on all of the identified operating variables, while the Power model is defined on a subset of the identified variables, excluding FGTI, FGT, and PHGOT. The problem is structured as a bi-level optimization where the upper level seeks to maximize power generation, while the lower level minimizes THR subject to operational constraints. The dataset is taken from \cite{ashraf2024driving}, and 579 samples are randomly divided into a training set (70$\%$) for model parameter learning and a validation set (30$\%$) for hyperparameter tuning and early stopping to mitigate over-fitting. Whereas, an independent testing set of 145 samples is reserved for final performance assessment of trained models.

Figure \ref{fig:data_surface} shows the parallel-plot made on the operating and performance variables of gas turbine system. The historical operation data reflect the noisy and asynchronous operation control schemes, heterogenous and complex interactions between the variables as evident from  Figure \ref{fig:data_surface}. This confirms that function of performance variables (Power and THR) built with operating variables is nonlinear and non-convex (refer to appendix \ref{vis_conv} to compare the parallel plot with those of benchmark problems). This, in turn, identifies nature of bi-level optimization problem (non-convex at both levels) which is defined and solved in the following.    

\begin{figure}[H]
    \centering
    \includegraphics[width=0.9\linewidth]{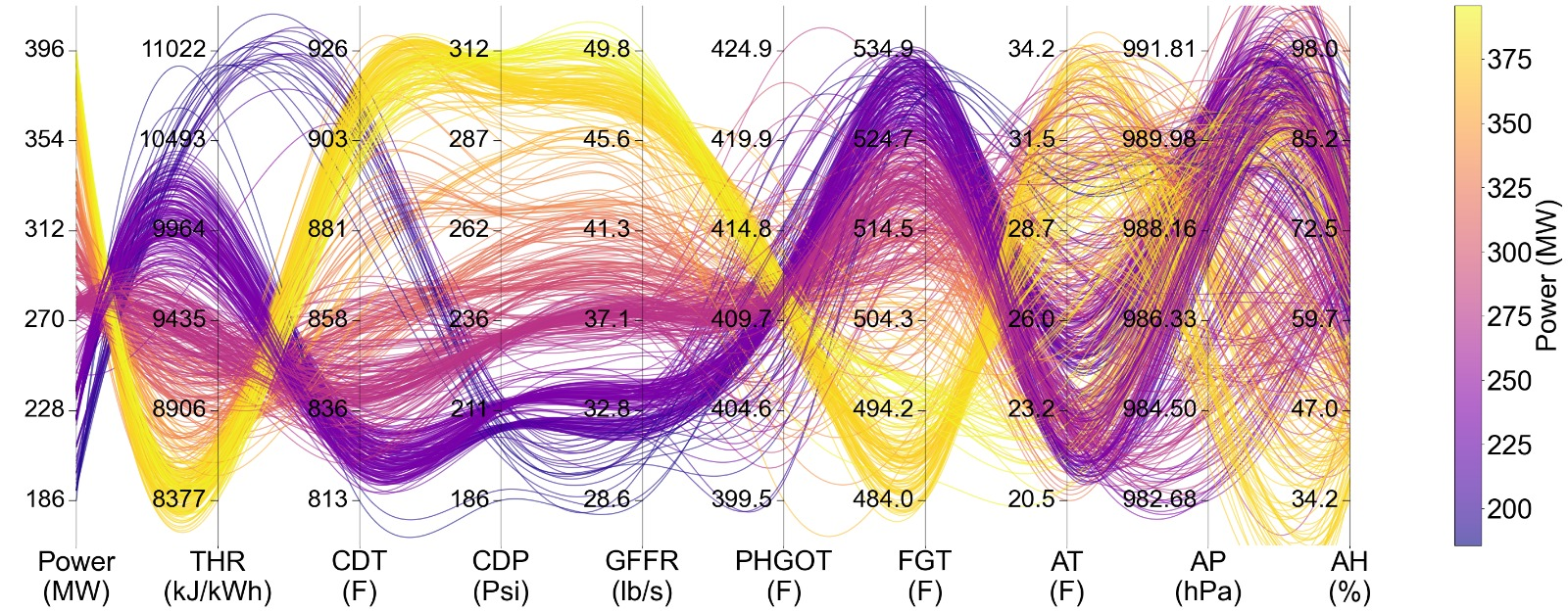}
    \caption{Parallel coordinates plot visualizing the multidimensional operating space of a 395 MW gas turbine system. The visualization highlights complex non-linear mapping network between the variables and non-convex function space of performance variables (Power and THR) built with the operating variables.}
    \label{fig:data_surface}
\end{figure}

\paragraph{Bi-level Problem Formulation}

The bi-level optimization problem for the gas turbine system is formulated as:

\begin{align}
\intertext{{Upper-level problem:}}
& \max_{x}
& & \hat{F}_{\text{ANN\_Power}}(x) \\
& \text{subject to}
& & x \in [0,1]^6, \quad x \in y,\notag \\
\intertext{{Lower-level problem:}}
& \min_{y}
& & \hat{f}_{\text{ANN\_THR}}(y) \notag \\
& \text{subject to}
& & (y - \mu)^T \Sigma^{-1} (y - \mu) \leq \tau^2, \label{eq:gt_lower_maha} \\
& & & y \in [0,1]^9. \notag
\end{align}

Here, $\hat{F}_{\text{ANN\_Power}}(x)$ represents the ANN surrogate model for power output (MW), while $\hat{f}_{\text{ANN\_THR}}(y)$ represents the ANN surrogate model for THR (kJ/kWh). The Mahalanobis distance-based constraint for lower-level problem is also embedded for estimating historical operation compliant computation of optimal solution.

Figure \ref{fig:bilevel_solutions} summarizes the numerical behavior of the bi-level optimization across different tolerance levels. Figure \ref{fig:bilevel_solutions}(a) highlights clear differences in solver trajectories, where a subset of runs converges to optimal solutions with stable objective values, while others exhibit oscillatory or stalled behavior accompanied by persistent constraint violations. This contrast indicates that feasibility is highly sensitive to the chosen tolerance level, which governs whether the solver can simultaneously satisfy constraints and improve the objective. Figure \ref{fig:bilevel_solutions} (b) visualizes the feasible operating envelope in terms of gas fuel flow rate and power output, with turbine heat rate represented through color contours. A number of optimal solutions are obtained on different $\tau$ levels, with the maximum power of 402 MW and turbine heat rate of 8159 kJ/kWh at $\tau = 89\%$. The obtained solutions follow a narrow manifold, illustrating the relationship between increasing power and maintaining low heat rate. Finally, Figure \ref{fig:bilevel_solutions} (c) shows that computational effort varies across tolerance levels, remaining less than one second to the solution convergence (optimal or infeasible). The CPU time consumbed for the identified optimal solution is 0.409 second. This is the key advantage of ANN-KKT framework to estimate optimal solutions in reasonable timeframe that can support process dynamics for variable power ramp-up and ramp-down operation yet minimizing turbine heat rate for the operation of gas turbine system.  

\begin{figure}[H]
    \centering
    \includegraphics[width=0.7\linewidth]{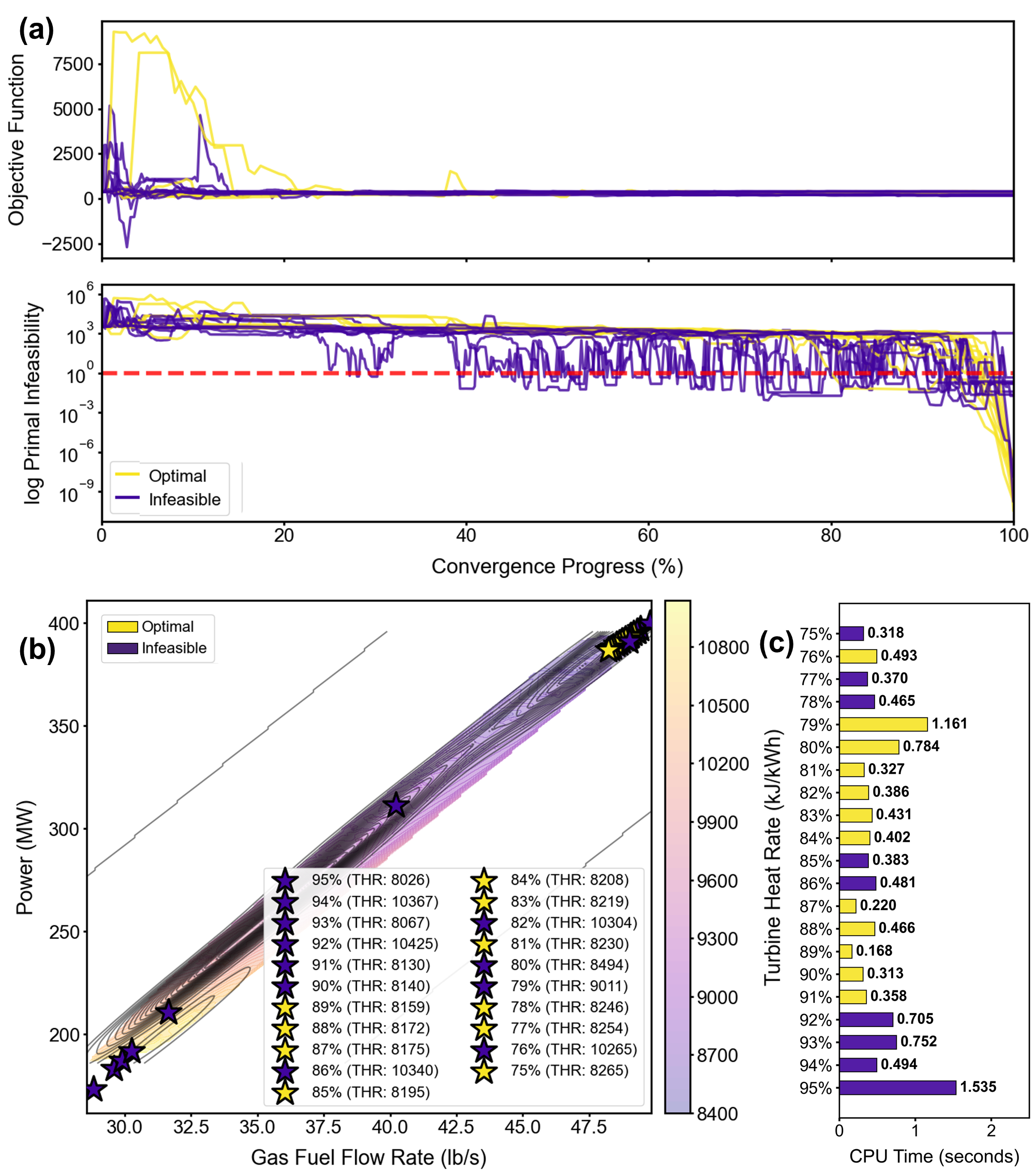}
    \caption{Optimal operating points from bi-level optimization of gas turbine system across Mahalanobis distance tolerances. (a) IPOPT convergence profiles for objective function and primal infeasibility constraint for bi-level optimization across tolerance levels. Primal infeasibility trajectories on logarithmic scale, with feasible solutions achieving constraint satisfaction below $10^{-6}$ and infeasible solutions stagnating above $10^0$ (red dashed reference line). (b) The surface shows turbine heat rate as a function of gas fuel flow rate and power output. Star markers indicate solutions computed for $\tau$ levels from 75\% to 95\%, with yellow stars representing feasible solutions and purple stars correspond to infeasible solutions. Black contour lines show historical operational density. (c) CPU time consumed to compute solutions corresponding to different values of $\tau$ for Mahalanobis constraint.}
    \label{fig:bilevel_solutions}
\end{figure}

\subsection{Robust Optimization through KKT formulation for Gas Turbine System}

ANN-KKT-based bi-level framework is extended to the robust optimization for maximising thermal efficiency (TE) of power generation operation of 395 MW gas turbine system. The problem is structured as a bi-level optimization where the upper level maximizes the stability radius ($\rho$) to ensure operational robustness for the computed solution. The lower level acts as an "adversary", minimizing TE within an uncertainty set defined by $\rho$ to identify the worst-case $\delta$. The robust optimization approach identifies the largest value of $\delta$ where TE remains higher than the target value ($TE_{target}$) for a given performance budget \cite{berthold2024unified, wiebe2022robust}. The robust optimization problem for maximizing TE above the $TE_{target}$ is formulated as follows:

\begin{align}
\intertext{{upper-level problem:}}
& \max_{x, \rho}
& & \rho \notag \\
& \text{subject to}
& & (x - \mu)^T \Sigma^{-1} (x - \mu) \leq {\tau}^2, \notag \\
& & & \left[ \min_{\delta} \hat{f}_{\text{ANN}}(x + \delta) \right] \geq TE_{\text{target}}, \label{eq:robust_safety_cond} \\
& & & x \in [0,1]^9, \quad x = [x_1, x_2,\dots,x_9] \notag \\
\intertext{{lower-level problem (Adversarial):}}
& \min_{\delta}
& & \hat{f}_{\text{ANN}}(x + \delta) \label{eq:uncer_set} \\
& \text{subject to}
& & \delta^T \Sigma^{-1} \delta \leq \rho^2 \label{sim_exp}
\end{align}

The ellipsoid uncertainty set ($\delta^T \Sigma^{-1} \delta \leq \rho^2$) is defined using the inverse covariance matrix ($\Sigma^{-1}$) derived from the training data ($\mu$). ${\tau}$ adjusts the ellipsoidal volume around the mean of multivariate joint-distribution and ellipsoid constraint $(x - \mu)^T \Sigma^{-1} (x - \mu)$ embeds the correlation structures which are respected while estimating the optimal solution. Similarly, identification of $\delta$ is also made while observing the correlation structures through ellipsoid constraint ($\delta^T \Sigma^{-1} \delta$) \cite{shang2017datadriven}. 

\begin{figure}[H]
    \centering
    \includegraphics[width=0.5\textwidth]{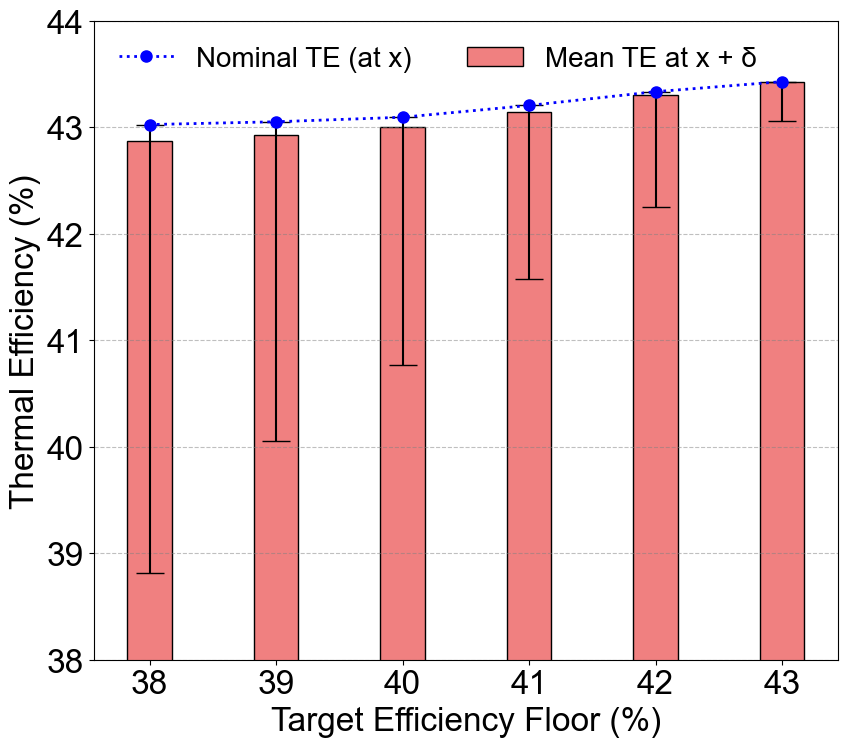} 
    \caption{Comparison of Thermal Efficiency on Nominal ($x$, blue dotted line) and Mean Robust ($x+\delta$, red bars) conditions under varying target efficiency floors ($TE_{target}$). The error bars represent the range of efficiency outcomes within the stability radius ($\rho$). Stricter efficiency targets force a reduction in $\rho$, resulting in smaller error bars and tighter convergence between nominal and robust operations.}
    \label{fig:compare}
\end{figure}

The ANN-KKT reformulation of robust optimization problem is mentioned in appendix \ref{rob_kkt}. The solution results of the ANN-KKT reformulation on different values of $TE_{target}$ are illustrated in Figure \ref{fig:compare}. $TE_{target}$ is labelled as Target Efficiency Floor on Figure \ref{fig:compare}. The blue dotted line represents the nominal TE at the optimal solution $x$, while the red bars represent the mean TE under worst-case perturbations ($x+\delta$). The operating ranges for each process variable based on computed value of $\delta$ under different $TE_{target}$ are mentioned in Table \ref{tab:robust_solutions}. For the cases where $\delta$ is computed to be negative than the nominal solution, the upper edge of operating ranges, mentioned in Table \ref{tab:robust_solutions}, are lower than those of lower edges.

\begin{figure}[H]
    \centering
    \includegraphics[width=1\textwidth]{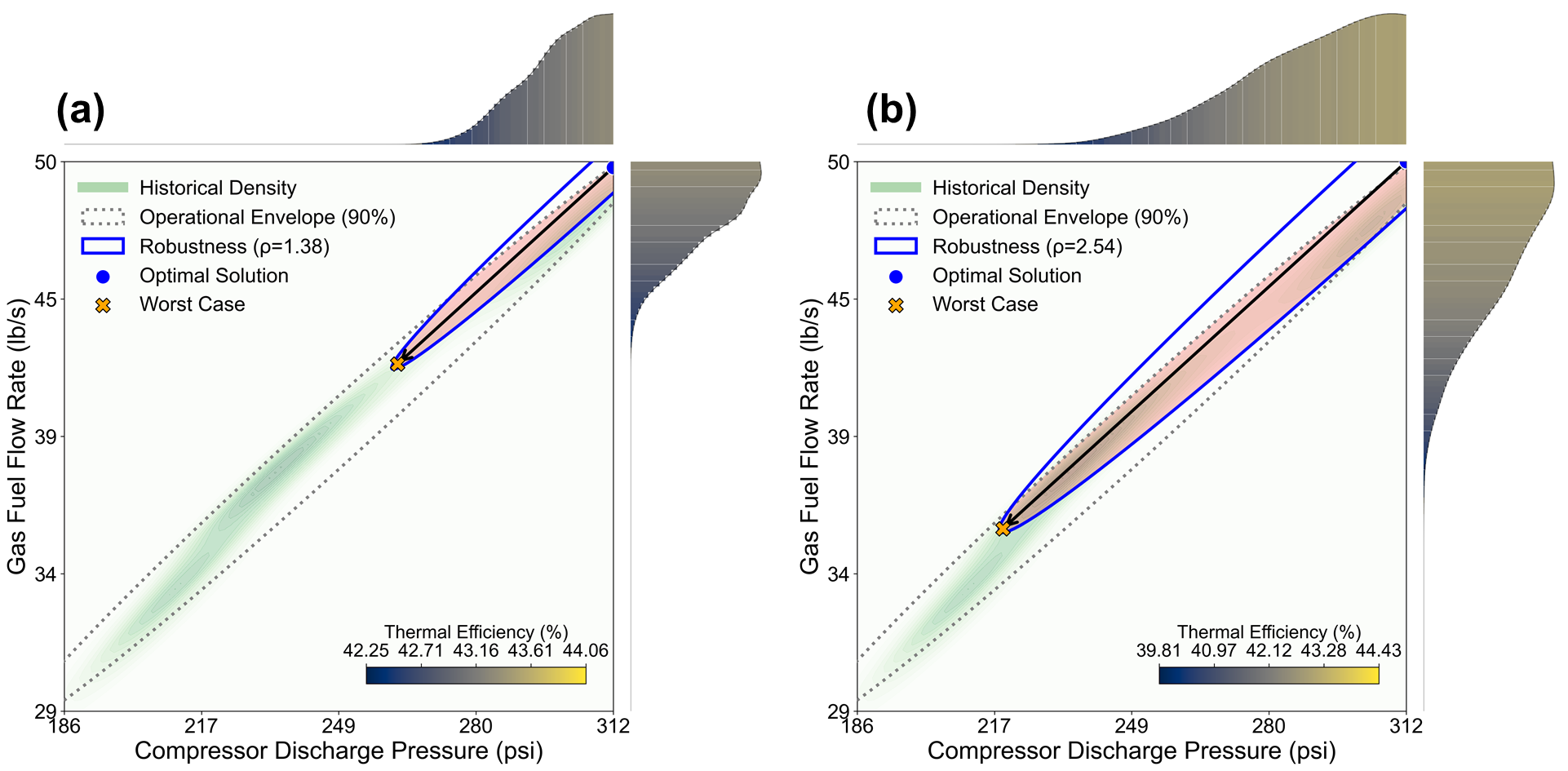} 
    \caption{Robustness landscape for operation of gas turbine system. (a) $TE_{target}$ = 42\%: A strict efficiency constraint results in a conservative stability radius ($\rho=1.38$), represented by the narrow blue ellipse. (b) $TE_{target}$ = 39\%: Lowering the target efficiency floor enhances the stability radius ($\rho=2.54$), as shown by the expanded blue ellipse. The green shading region represents the density of historical operating data and dotted ellipse is constructed on 90$^{th}$ quantile. The pink-colored shaded region represents the feasible operating space where TE remains higher than $TE_{target}$.}
    \label{fig:robust_landscape}
\end{figure}

10,000 perturbations are generated within the operating space of variables defined on the computed value of $\delta$ and the simulated experiments follow the correlation structure of variables (refer to \ref{sim_exp}). The designed computational experiments (x + $\delta$), having operating levels of variables outside the Mahalanobis-based operating envelop (90$^{th}$), are discarded (refer to Figure \ref{fig:robust_landscape} for visualization). The error bars on mean TE in Figure \ref{fig:compare} indicate the spread in TE against the computational experiments (x + $\delta$). It is observed that as the strictness of the Target Efficiency Floor increases (from 38\% to 43\%), the gap between nominal TE and Mean TE (computed on $x+\delta$) decreases. This indicates the lower value of $\rho$, meaning shorter stability radius where TE remains feasible under perturbations in optimal solution ($x$). This can also be attributed to upper limit in TE that can be achieved from power generation operation of gas turbine system and the process control space near the upper limit is small for TE to remain robust to perturbations in process conditions ($x+\delta$).

\begin{table}[H]
\centering
\caption{Optimal operating conditions under varying target efficiency floors ($TE_{target}$). For each efficiency target, $[x, x+\delta]$ represents the operating range where $x$ is the nominal physical operating point and $x+\delta$ is the worst-case physical value under maximum allowable perturbations within the stability radius ($\rho$).}
\label{tab:robust_solutions}
\small
\begin{tabular}{lcccccc}
\toprule
\textbf{Feature} & \textbf{38\%} & \textbf{39\%} & \textbf{40\%} & \textbf{41\%} & \textbf{42\%} & \textbf{43\%} \\
\midrule
AT & [29.63, 25.05] & [29.82, 25.71] & [30.11, 26.51] & [29.84, 26.86] & [29.50, 27.33] & [29.01, 28.17] \\
AP & [987, 988] & [987, 988] & [987, 988] & [988, 988] & [988, 989] & [989, 989] \\
AH & [50.22, 65.25] & [49.46, 63.00] & [48.48, 60.36] & [49.38, 59.30] & [50.71, 58.07] & [53.09, 56.10] \\
CDP & [312.0, 209.0] & [312.0, 219.3] & [312.0, 230.8] & [312.0, 244.5] & [312.0, 262.4] & [312.0, 292.4] \\
GFFR & [49.83, 34.04] & [49.83, 35.62] & [49.83, 37.38] & [49.77, 39.41] & [49.61, 42.00] & [49.37, 46.35] \\
CDT & [861, 783] & [861, 791] & [861, 801] & [860, 811] & [860, 825] & [864, 852] \\
FGEXT & [640, 680] & [640, 676] & [640, 672] & [639, 665] & [637, 657] & [635, 643] \\
FGT & [484, 525] & [484, 521] & [484, 516] & [485, 512] & [487, 506] & [489, 497] \\
PHGOT & [412.5, 412.9] & [412.6, 413.0] & [412.8, 413.1] & [413.2, 413.3] & [413.6, 413.6] & [414.1, 414.0] \\
\bottomrule
\end{tabular}
\end{table}

The robustness of computed solutions on different values of $TE_{target}$ is graphically inspected that satisfies the operational constraints of the power generation process. The data-driven operational regimes are constructed for two operating variables, namely compressor discharge pressure (psi) and gas fuel flow rate (lb/s) and the operational envelop is drawn to capture 90\% of the operational regime (shown by dotted ellipse) on Figure \ref{fig:robust_landscape}. The stability radius ($\rho$) computed from robust optimization is plotted by blue colored ellipse that corresponds to worst-cases ($\rho = 1.38$, $\rho = 2.54$) for identifying the feasible operating space of the variables (pink-colored shaded region) to maintain TE above $TE_{target}$ of 42\% and 39\%, respectively. The identified operating space hedges the uncertainty in the operating variables and can maintain TE above the set value. The distribution of TE computed on the simulated experiments (x+$\delta$) and corresponding to identified operating space is shown on the top edges on Figure \ref{fig:robust_landscape}. It is evident from the distribution that TE is computed to higher than $TE_{target}$ of 42\% and 39\% corresponding to identified operating space.

The ANN-KKT framework demonstrates the suitability for data-driven optimization of thermal power systems and scales up to robust optimization for the robust identification of operating space to hedge the operational uncertainty. The domain-compliant AI-powered analytics can assist the existing control schemes implemented in industrial thermal power systems to enhance the efficiency of industrial processes. The human-AI collaboration for the implementation of ANN-KKT framework (injecting historical information in analytics through setting up $\tau$) fosters reliability and constraints satisfaction that enhances AI access to process control of industrial systems. This is aligned with the recommendations of International Energy Agency to design domain-compliant AI systems which improve the process control to harvest higher energy efficiency to support industrial decarbonization \cite{ieaai}.

\section{Conclusion}

The scaleable and computationally efficient solution computation methodology for competing objectives of bi-level problems are key challenges to scale its adoption to industrial power systems. In this paper, we present ANN-KKT framework that embeds ANN surrogate models in the objective function at both levels of bi-level framework and reformulate it into single level problem by replacing the lower-level problem with KKT conditions. The ANN-KKT framework provides comparable solutions with the bi-level solutions of the benchmark problems. When applied to maximizing the power generation and minimizing the turbine heat rate for thermal power plants, the ANN-KKT framework provides optimal solutions (maximum power generation of 583 MW and 402 MW \& turbine heat rate of 7337 kJ/kWh and 7542 kJ/kWh for coal and gas turbine system, respectively) with the computational time ranging from 0.22 s to 0.88 s. Additionally, ANN-KKT framework is also able to identify robust operating space for the process variables, hedging the process uncertainty to achieve thermal efficiency higher than 42\% for gas turbine system. These results demonstrate the scalability and computational efficient performance of ANN-KKT framework to compute optimal and domain-consistent solutions for different nature of industrial operations (maximizing power generation on minimum heat rate, ensuring high thermal efficiency against the process uncertainty) for thermal power systems. The advances in the development of computational frameworks for enhancing the efficiency of industrial power generation processes contribute to net-zero mission and industry 5.0.  

\section{Limitations and Future Work}
We have embedded Mahalanobis distance-based constraint that adjust the search space for the optimization solver based on the value of $\tau$. The tuning of $\tau$ is challenging and it affects the feasibility of solutions, rendering it one of the limitations of this study. ANN model introduce non-convexity in the optimization problem and may complicate the solution convergence profiles; future research may consider convexification of ANN models that may expedite the numerical convergence performance of the optimization solver. 

\section*{Acknowledgement}
Waqar Muhammad Ashraf acknowledges that this work was supported by the Alan Turing Institute’s Enrichment Scheme/Turing Studentship Scheme.

\section*{Author Contributions}
\textbf{Talha Ansar:} Methodology, Data Curation, Software,  Validation, Formal Analysis, Writing – Original Draft, \textbf{Muhammad Mujtaba Abbas:} Validation, Writing – Review \& Editing, \textbf{Ramit Debnath:} Validation, Writing – Review \& Editing, \textbf{Vivek Dua:} Validation, Writing – Review \& Editing,  \textbf{Waqar Muhammad Ashraf:} Conceptualization, Methodology, Data Curation, Software,  Writing – Original Draft, Project Administration

\section*{Data availability statement}
The code and data can be provided on request.

\section*{Declaration of competing interests}
The authors declare no competing interests.

\printbibliography

\newpage
\begin{appendices}
\section*{Appendix}
\section{Neural Network Architecture and Training} \label{sec:modelling}

ANN models are constructed using the PyTorch framework.  The Sigmoid Linear Unit (SiLU) activation function is applied on the hidden layer while linear activation function is implemented on the output layer of ANN to ensure the smooth, continuously differentiable gradients required for subsequent KKT-based reformulation. Leveraging the capability of Hyperopt to handle asynchronous distribution spaces, 50 optimization trials were performed per model to minimize validation RMSE. The search space comprised a uniform integer distribution for hidden neurons ($N_h \in [2, 16]$) and log-uniform distributions for the learning rate ($[10^{-5}, 10^{-1}]$), $\lambda_1$ regularization ($[10^{-7}, 10^{-1}]$), and $\lambda_2$ ($[10^{-7}, 10^{-1}]$). Finally, training episodes were extended to a maximum of 5,000 epochs, governed by an early stopping protocol with a patience of 200 epochs. This mechanism ensured training termination upon performance saturation, automatically restoring the model parameters associated with the lowest validation RMSE to guarantee optimal generalization.

\section{Benchmark Problems} \label{sec:bench_extended}

\subsection{Visualization of convex and con-convex functions} \label{vis_conv}
The topological landscapes of the objective functions for the three benchmark problems C \& C, C \& NC and NC \& NC are visualized below in Figures \ref{fig:vis_cc}, \ref{fig:vis_cnc} and \ref{fig:vis_ncnc} respectively. Each figure comprises: (a) a parallel coordinates plot mapping the interactions between decision variables ($x, y$) and objective values; (b) a 3D surface plot of the upper-level objective function $F(x, y)$; and (c) a 3D surface plot of the lower-level objective function $f(x, y)$.

\begin{figure}[H]
    \centering
    \includegraphics[width=0.85\textwidth]{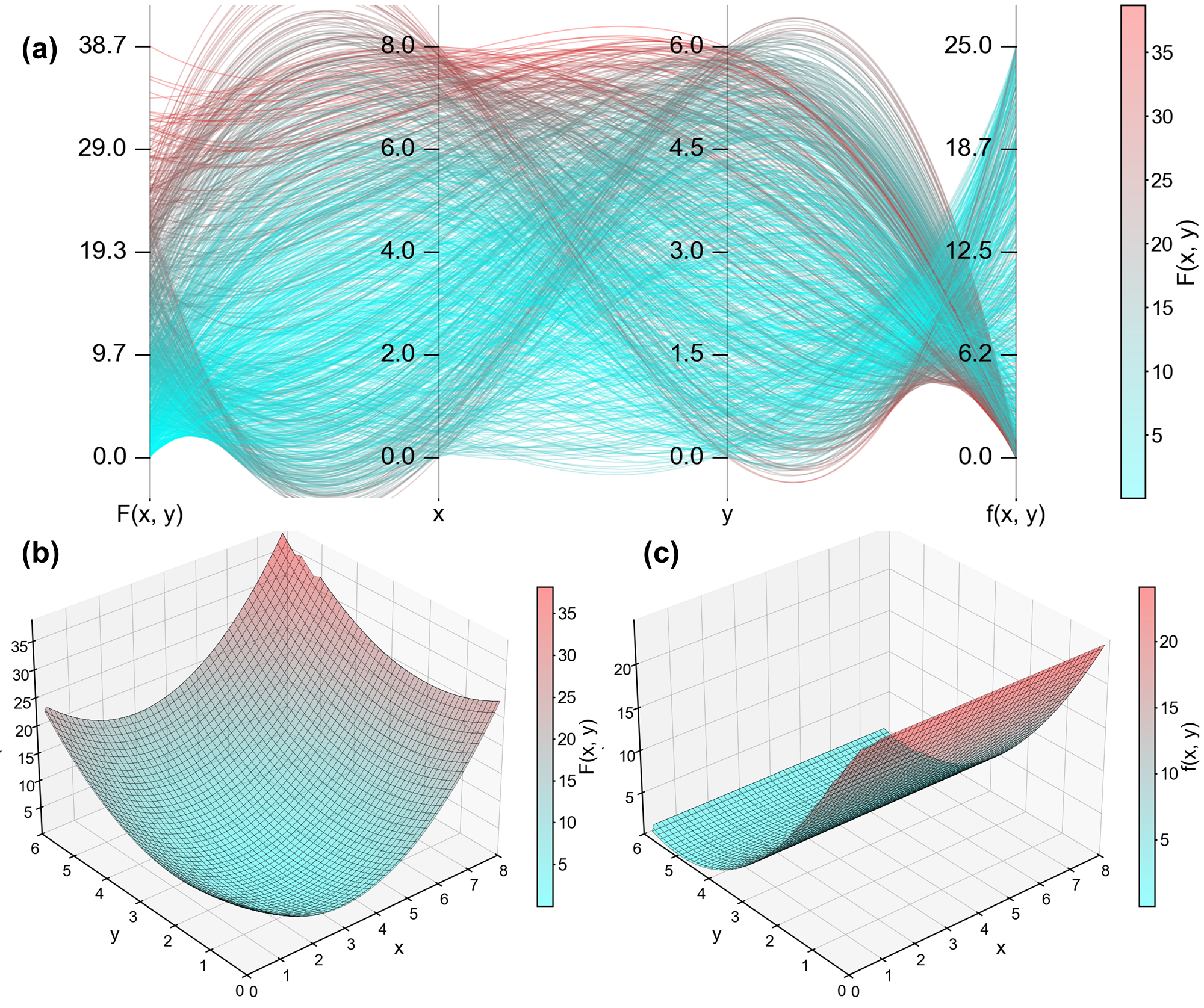}
    \caption{Convex \& Convex: (b) and (c) display convex, bowl-shaped surfaces for both upper and lower level objectives, facilitating stable gradient descent and global optimality.}
    \label{fig:vis_cc}
\end{figure}

\begin{figure}[H]
    \centering
    \includegraphics[width=0.85\textwidth]{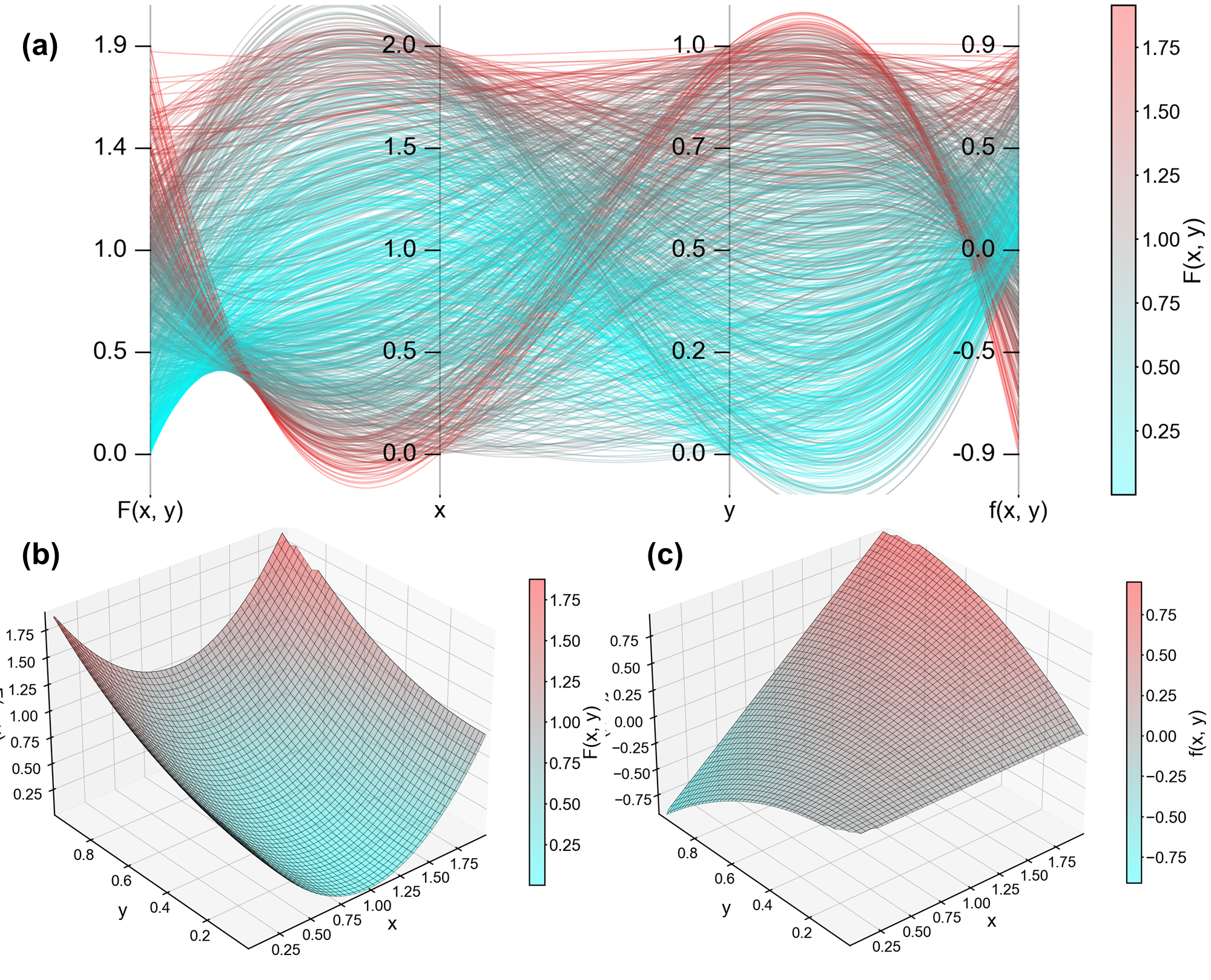}
    \caption{Convex \& Non-Convex: While the upper-level objective (b) remains convex, the lower-level objective (c) exhibits a saddle-point geometry (indefinite quadratic), introducing non-convexity into the KKT reformulation.}
    \label{fig:vis_cnc}
\end{figure}

\begin{figure}[H]
    \centering
    \includegraphics[width=0.85\textwidth]{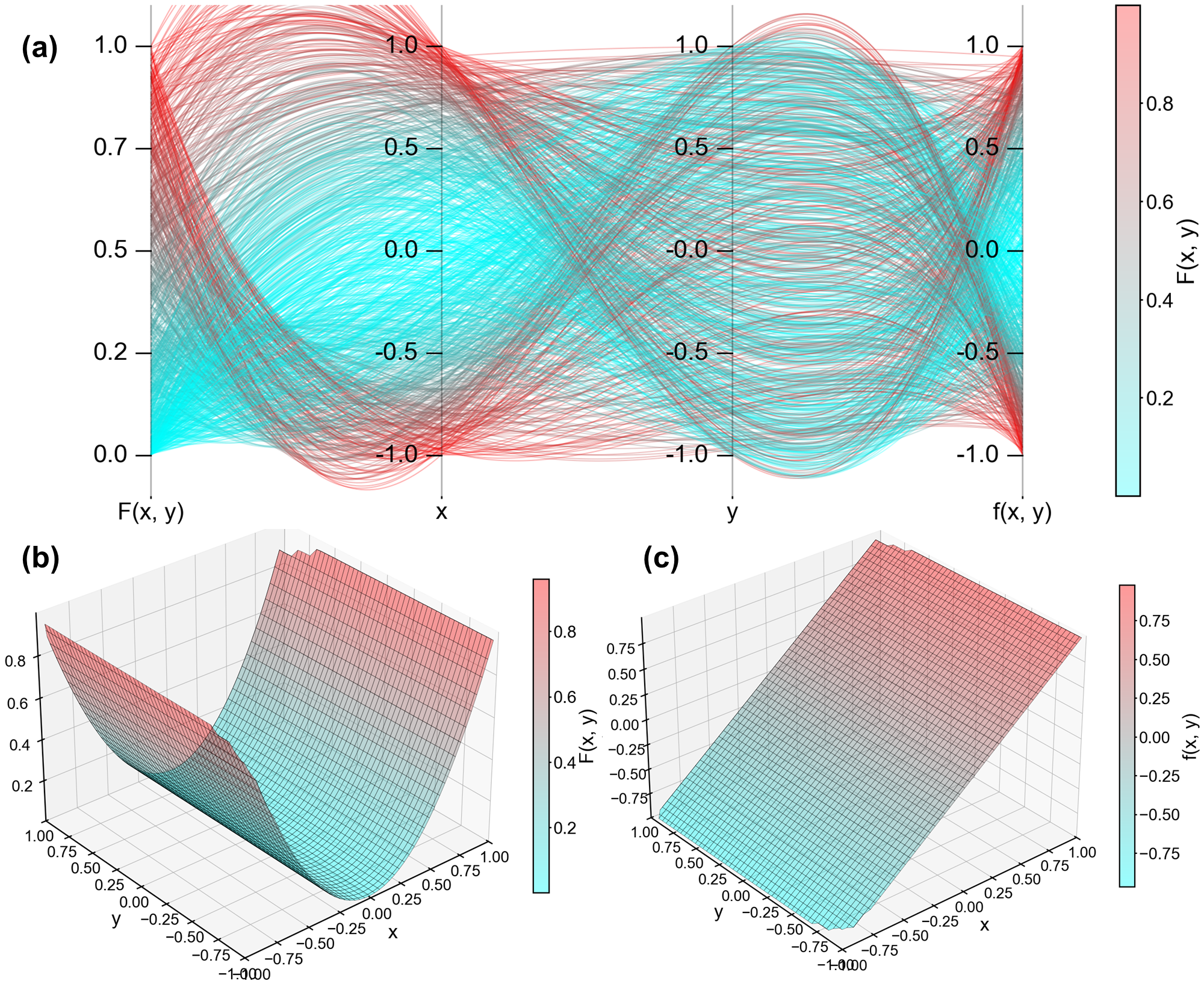}
    \caption{Non-Convex \& Non-Convex: The upper level (b) forms a parabolic cylinder and the lower level (c) a linear plane. The problem's non-convexity and difficulty primarily arise from the non-convex constraints restricting these surfaces.}
    \label{fig:vis_ncnc}
\end{figure}

\subsection{Training of ANN models and performance metrics}
The final models are evaluated using Root Mean Squared Error (RMSE) and the coefficient of determination ($R^2$) on the hold-out test set to ensure the development of high-fidelity surrogate models for the bi-level optimization task. Table \ref{tab:ann_performance} shows that the ANN surrogates achieve low RMSE and consistently high $R^2$ values across training, validation, and test sets for all benchmark problems. The close agreement of metrics across data splits indicates good generalization. These results confirm the suitability of the trained models for integration into the bi-level optimization framework.

\begin{table}[ht]
\caption{Performance Metrics of ANN Surrogates for Benchmark Bi-level Optimization Problems}
\label{tab:ann_performance}
\centering
\begin{tabular}{llcccccc}
\toprule
Problem Type & Level & \multicolumn{2}{c}{Train} & \multicolumn{2}{c}{Validation} & \multicolumn{2}{c}{Test} \\
\cmidrule(lr){3-4} \cmidrule(lr){5-6} \cmidrule(lr){7-8}
 &  & RMSE & $R^2$ & RMSE & $R^2$ & RMSE & $R^2$ \\
\midrule
Convex \& & Upper & $3.74\times10^{-2}$ & 0.99 & $3.99\times10^{-2}$ & 0.99 & $4.24\times10^{-2}$ & 0.99 \\
Convex & Lower & $3.67\times10^{-2}$ & 0.99 & $3.73\times10^{-2}$ & 0.99 & $3.80\times10^{-2}$ & 0.99 \\
\midrule
Convex \& & Upper & $4.72\times10^{-3}$ & 0.99 & $4.78\times10^{-3}$ & 0.99 & $4.84\times10^{-3}$ & 0.99 \\
Nonconvex & Lower & $3.62\times10^{-3}$ & 0.99 & $3.88\times10^{-3}$ & 0.99 & $4.13\times10^{-3}$ & 0.99 \\
\midrule
Nonconvex \& & Upper & $6.47\times10^{-4}$ & 0.99 & $6.96\times10^{-4}$ & 0.99 & $7.44\times10^{-4}$ & 0.99 \\
Nonconvex & Lower & $2.58\times10^{-4}$ & 0.99 & $2.61\times10^{-4}$ & 0.99 & $2.64\times10^{-4}$ & 0.99 \\
\bottomrule
\end{tabular}
\end{table}

\subsection{MPEC Reformulations for Additional Benchmark Problems}

We provide the ANN-based single-level reformulations for the remaining two benchmark problems (Convex \& Non-convex and Non-convex \& Non-convex) using KKT conditions. Consistent with the methodology applied to the SC2 problem, the objective functions at both levels are replaced by ANN surrogates, while the constraints remain explicit.

\subsubsection{Convex \& Non-convex} \label{ANN_KKT_B1}

This problem features a convex upper-level and a non-convex lower-level objective with box constraints.

Objective functions:
\begin{equation}
\begin{gathered}
    \min_{x, y, \boldsymbol{\mu}} \quad \hat{F}_{\text{ANN}}(x,y) \\
    \text{Here }\qquad \hat{F}_{\text{ANN}}(x,y) \approx (x-1)^2 + y^2 \\
     \qquad \hat{f}_{\text{ANN}}(x,y) \approx -y^2 + xy
\end{gathered}
\end{equation}

Upper-Level Constraints:
\begin{equation}
x \in [0, 2] 
\end{equation}

Stationarity Condition:
\begin{equation}
\nabla_{y} \hat{f}_{\text{ANN}}(x, y) - \mu^L + \mu^U = 0 
\end{equation}

where $\nabla_{y} \hat{f}_{\text{ANN}}(x, y)$ is computed via symbolic differentiation of the trained ANN model. $\mu^L$ and $\mu^U$ are the multipliers for the lower ($0$) and upper ($1$) bounds of $y$, respectively.

Primal Feasibility:
\begin{align}
y &\in [0, 1] 
\end{align}

Dual Feasibility:
\begin{align}
\mu^L, \mu^U &\geq 0 
\end{align}

Complementarity Conditions:
\begin{align}
\mu^L \cdot -y &= 0 \\
\mu^U \cdot (y - 1) &= 0 
\end{align}

\subsubsection{Non-convex \& Non-convex } \label{ANN_KKT_B2}

The "Pathological Branching" problem includes a difficult non-convex constraint in the lower level ($y^2(x - 0.5) \leq 0$) and non-convex objectives.

Objective functions:
\begin{equation}
\begin{gathered}
    \min_{x, y, \lambda_1, \boldsymbol{\mu}} \quad \hat{F}_{\text{ANN}}(x,y) \\
    \text{Here }\qquad \hat{F}_{\text{ANN}}(x,y) \approx x^2 \\
     \qquad \hat{f}_{\text{ANN}}(x,y) \approx y
\end{gathered}
\end{equation}

Upper-Level Constraints:
\begin{align}
1 + x - 9x^2 - y &\leq 0 \\
x &\in [-1, 1]
\end{align}

Stationarity Condition:
\begin{equation}
\nabla_{y} \hat{f}_{\text{ANN}}(y) + \lambda_1 \nabla_{y} (y^2(x - 0.5)) - \nabla_{y_i} (\mu_i^L . (-y-1) + \mu_i^U . (y-1)) = 0 
\end{equation}

The gradient of the non-convex lower-level constraint is computed explicitly as $\nabla_{y} (y^2(x - 0.5)) = 2y(x - 0.5)$.

Primal Feasibility:
\begin{align}
y^2(x - 0.5) &\leq 0 \\
y &\in [-1, 1] 
\end{align}

Dual Feasibility:
\begin{align}
\lambda_1 &\geq 0 \\
\mu^L, \mu^U &\geq 0 
\end{align}

Complementarity Conditions:
\begin{align}
\lambda_1 \cdot (y^2(x - 0.5)) &= 0 \\
\mu^L \cdot (- y - 1) &= 0 \\
\mu^U \cdot (y - 1) &= 0 
\end{align}

\subsection{Local Solver Solutions of Benchmark Problems} \label{bench-prob}

\begin{table}[H]
\centering
\caption{Comparison of solutions and CPU times for solving different bi-level problems (Convex \& Convex, Convex \& Non-Convex, Non-Convex \& Non-Convex). Book/bi-level solutions are from the cited test sets; BARON and IPOPT results are reported for Bi-level-KKT and ANN-KKT} reformulations.
\label{tab:merged_results_all}
\begin{tabular}{llcccc}
\toprule
\textbf{Problem Type} & \textbf{Method} & \textbf{x} & \textbf{y} & \textbf{Objective (F)} & \textbf{Time (s)} \\
\midrule

\multirow{5}{*}{C \& C} 
 & Bi-level \cite{mitsos2006testset} & 1 & 3 & 5 & - \\
 & Bi-level-KKT (BARON) & 1 & 3 & 5 & 0.14 \\
 & ANN-KKT (BARON) & 0.9810 & 2.9620 & 4.9760 & 1.27 \\
 & Bi-level-KKT (IPOPT) & 4.4 & 4.8 & 9.8 & 0.022 \\
 & ANN-KKT (IPOPT) & 3.0179 & 4.9342 & 12.98 & 0.013 \\
\cmidrule{1-6}

\multirow{5}{*}{C \& NC} 
 & Bi-level \cite{mitsos2008global} & 1 & 0 & 0 & - \\
 & Bi-level-KKT (BARON) & 1 & 0 & 0 & 0.12 \\
 & ANN-KKT (BARON) & 1.0140 & 0 & 0.0034 & 1.27 \\
 & Bi-level-KKT (IPOPT) & 0.8 & 0.4 & 0.2 & 0.009 \\
 & ANN-KKT (IPOPT) & 1.0144 & 0 & 0.0034 & 0.036 \\
\cmidrule{1-6}

\multirow{5}{*}{NC \& NC} 
 & Bi-level \cite{mitsos2006testset} & -0.4191 & -1 & 0.1756 & - \\
 & Bi-level-KKT (BARON) & -0.4191 & -1 & 0.1756 & 0.09 \\
 & ANN-KKT (BARON) & -0.4191 & -1 & 0.1750 & 0.39 \\
 & Bi-level-KKT (IPOPT) & -0.4191 & -1 & 0.1757 & 0.009 \\
 & ANN-KKT (IPOPT) & -0.4191 & -1 & 0.1750 & 0.012 \\
\bottomrule
\end{tabular}
\end{table}

\begin{figure}[H]
    \centering
    \includegraphics[width=0.7\linewidth]{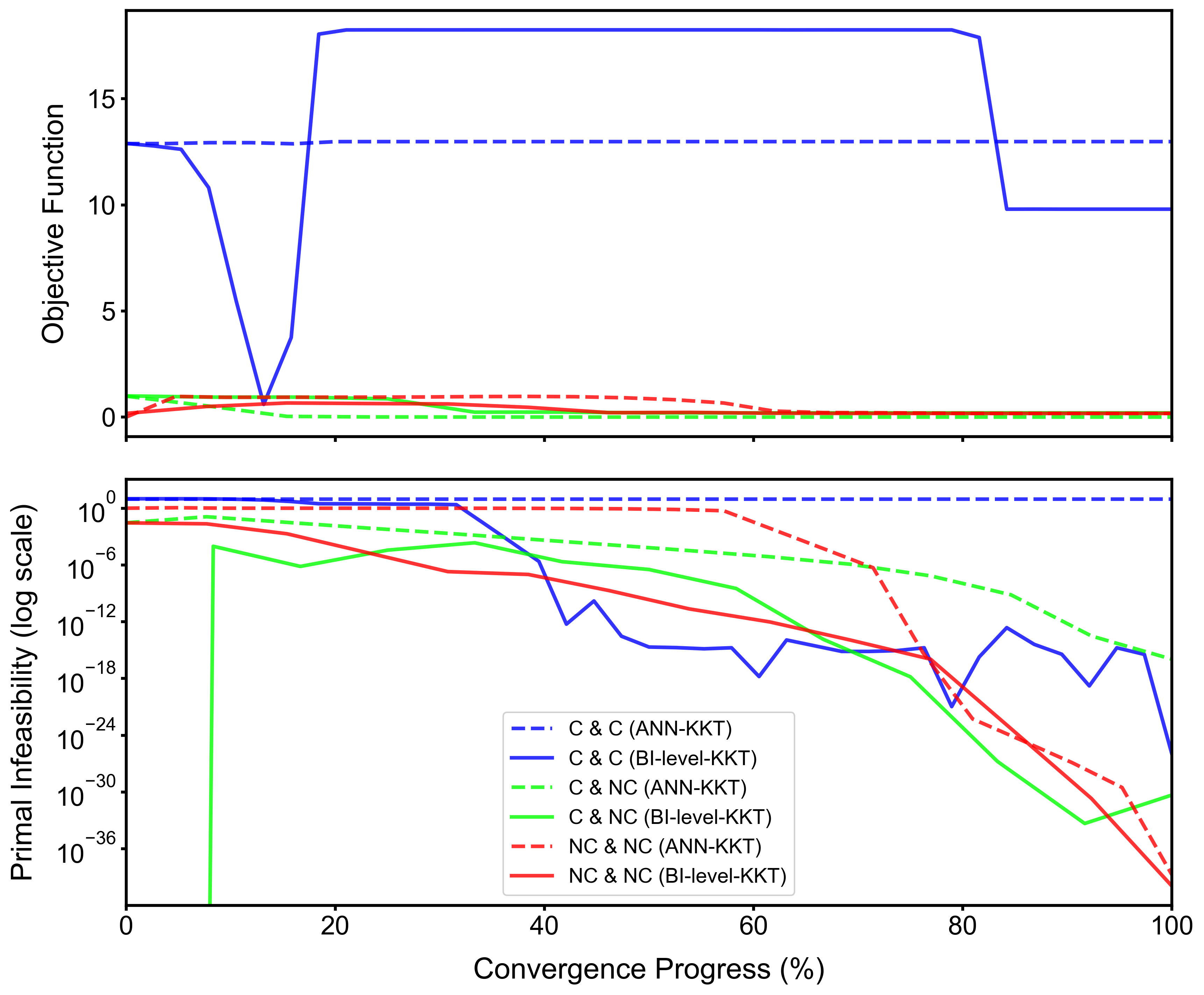}
    \caption{IPOPT convergence analysis for bi-level optimization of benchmark examples. Top: Convergence progress of solver for the objective function on ANN-KKT and Bi-level-KKT reformulations for three benchmark examples. Bottom: The profiles of primal infeasibility against the convergence progress for the benchmark problems. The solution for C \& C problem remains infeasible as KKT constraints are not satisfied.}
    \label{fig:bench_convergence}
\end{figure}

\section{Bi-level Operation Optimization of Thermal Power Plants}

\subsection{660 MW coal power plant} \label{kkt_coal}

\subsubsection{Modelling performance}

The trained ANN models for Power and THR have 6 neurons in the hidden layer whereas optimized learning rate is found to be 0.06 and 0.058, respectively. The predictive performance of the trained models on training, validation and testing datasets is mentioned in Table \ref{tab:coal_performance}.  

\begin{table}[ht]
\centering
\caption{Performance metrics of ANN surrogates for coal power plant}
\label{tab:coal_performance}
\resizebox{\textwidth}{!}{%
\begin{tabular}{lcccccc}
\hline
\multirow{2}{*}{\textbf{Target Variable}} & \multicolumn{2}{c}{\textbf{Train}} & \multicolumn{2}{c}{\textbf{Validation}} & \multicolumn{2}{c}{\textbf{Test}} \\ \cline{2-7} 
 & \textbf{RMSE} & \textbf{$R^2$} & \textbf{RMSE} & \textbf{$R^2$} & \textbf{RMSE} & \textbf{$R^2$} \\ \hline
Power (MW) & 2.48 & 0.99 & 2.45 & 0.99 & 4.27 & 0.99 \\
Turbine Heat Rate (kJ/kWh) & 82 & 0.77 & 80 & 0.78 & 90 & 0.72 \\ \hline
\end{tabular}%
}
\end{table}

\subsubsection{ANN-KKT reformulation} \label{kkt_coal_plant}

The bi-level problem is reformulated as a single-level MPEC using KKT optimality conditions for the lower-level problem:

\begin{align}
& \max_{x, \mu^L, \mu^U, \lambda_m}
& & \hat{F}_{\text{ANN\_Power}}(x) \label{eq:tpp_mpec_obj}
\end{align}

Stationarity Conditions:

For each decision variable $y_i$, $i \in \{1,\ldots,8\}$:

\begin{equation}
\nabla_{y_i} \hat{f}_{\text{ANN\_THR}}(y) + \nabla_{y_i} (\mu_i^L . (-y) + \mu_i^U . (y-1))  + \lambda_m \cdot \nabla_{y_i} \left[(y - \mu)^T \Sigma^{-1} (y - \mu) - \tau^2\right] = 0 \label{eq:tpp_stationarity}
\end{equation}

The gradient of the ANN model $\nabla_{y_i} \hat{f}_{\text{ANN\_THR}}(y)$ and the gradient of the Mahalanobis distance constraint $\nabla_{y_i} \left[(y - \mu)^T \Sigma^{-1} (y - \mu) - \tau^2\right]$ are obtained through symbolic differentiation, while $- \mu_i^L$ and $\mu_i^U$ are the multipliers of the box constraints.

Highly non-convex function profile of objective functions built on ANN-based models of Power and THR (data-driven approximation and shown on \ref{fig:data_surface}) may result in numerical instability for the convergence performance. The primal Mahalanobis distance constraint (Eq. \ref{eq:coal_tpp_primal_maha}), its dual (Eq. \ref{eq:coal_tpp_dual_maha}) and complementarity (Eq. \ref{eq:coal_tpp_comp_maha}) are embedded without FB transformation. The complementarity conditions for box constraints are reformulated with $\Phi_{FB}^*$ to ensure numerical stability since constraint violations were observed without transformation. The reformulated complimentary constraints for box constraints (Eq. \ref{eq:coal_tpp_comp_lower} - Eq. \ref{eq:coal_tpp_comp_upper}) and inner-level constraints are given in the following:



\begin{align}
    \sqrt{(y_i)^2 + (\mu_i^L)^2 + \epsilon_l} - y_i - \mu_i^L &= 0, \quad \forall i, \label{eq:coal_tpp_comp_lower} \\
    \sqrt{(1 - y_i)^2 + (\mu_i^U)^2 + \epsilon_u} - (1 - y_i) - \mu_i^U &= 0, \quad \forall i, \label{eq:coal_tpp_comp_upper} \\[1em]
    (y - \mu)^T \Sigma^{-1} (y - \mu) - \tau^2 &\leq 0, \label{eq:coal_tpp_primal_maha} \\
    \lambda_m &\geq 0, \label{eq:coal_tpp_dual_maha} \\
    \left[(y - \mu)^T \Sigma^{-1} (y - \mu) - \tau^2\right] \cdot \lambda_m &= 0. \label{eq:coal_tpp_comp_maha}
\end{align}

where $\epsilon_l = 10^{-3}$ and $\epsilon_u = 10^{-9}$ are small regularization parameters to smooth the complementarity constraints. The reformulated MPEC is solved using the IPOPT solver through the GAMS interface in Pyomo.

\begin{table}[ht]
\centering
\caption{Optimality, CPU Time, and Solution Quality of the Coal Power Plant Optimization under Varying Tolerance Levels}
\label{tab:coal_solutions}
\begin{tabular}{ccccc}
\toprule
Tolerance (\%) & THR (kJ/kWh) & Power (MW) & Optimality & CPU Time (s) \\
\midrule
81 & 7423 & 551 & Optimal & 0.066 \\
82 & 7418 & 554 & Optimal & 0.186 \\
83 & 7772 & 386 & Infeasible & 0.682 \\
84 & 7407 & 423 & Optimal & 0.131 \\
85 & 7774 & 562 & Infeasible & 0.306 \\
86 & 7396 & 398 & Optimal & 0.053 \\
87 & 7391 & 535 & Optimal & 0.014 \\
88 & 7386 & 568 & Optimal & 0.015 \\
89 & 7380 & 570 & Optimal & 0.028 \\
90 & 7376 & 534 & Optimal & 0.018 \\
91 & 7369 & 574 & Optimal & 0.015 \\
92 & 7362 & 533 & Optimal & 0.016 \\
93 & 7343 & 533 & Optimal & 0.565 \\
94 & 7337 & 583 & Optimal & 0.052 \\
95 & 7766 & 533 & Infeasible & 0.564 \\
\bottomrule
\end{tabular}
\end{table}

\subsection{395 MW Gas power plant} \label{kkt_gas}

\subsubsection{Modelling performance}

The trained ANN models for Power and THR have 12 and 16 neurons in the hidden layer whereas optimized learning rate is found to be 0.090 and 0.098, respectively. The predictive performance of the trained models on training, validation and testing datasets is mentioned in Table \ref{tab:gas_performance}. 

\begin{table}[ht]
\centering
\caption{Performance Metrics of ANN Surrogates (Gas Turbine Power Plant)}
\label{tab:gas_performance}
\resizebox{\textwidth}{!}{%
\begin{tabular}{lcccccc}
\hline
\multirow{2}{*}{\textbf{Target Variable}} & \multicolumn{2}{c}{\textbf{Train}} & \multicolumn{2}{c}{\textbf{Validation}} & \multicolumn{2}{c}{\textbf{Test}} \\ \cline{2-7} 
 & \textbf{RMSE} & \textbf{$R^2$} & \textbf{RMSE} & \textbf{$R^2$} & \textbf{RMSE} & \textbf{$R^2$} \\ \hline
Power (MW) & 0.84 & 0.99 & 0.90 & 0.99 & 1.06 & 0.99 \\
Turbine Heat Rate (kJ/kWh) & 190 & 0.89 & 213 & 0.85 & 228 & 0.85 \\ \hline
\end{tabular}%
}
\end{table}

\subsubsection{ANN-KKT reformulation}

The bi-level problem is reformulated as a single-level MPEC using KKT optimality conditions for the lower-level problem:

\begin{align}
& \max_{x, \mu^L, \mu^U, \lambda_m}
& & \hat{F}_{\text{ANN\_Power}}(x) \label{eq:tpp_mpec_obj}
\end{align}

Stationarity Conditions:

For each decision variable $y_i$, $i \in \{1,\ldots,9\}$:

\begin{equation}
\nabla_{y_i} \hat{f}_{\text{ANN\_THR}}(y) - \nabla_{y_i} (\mu_i^L . y + \mu_i^U . (y-1))+ \lambda_m \cdot \nabla_{y_i} \left[(y - \mu)^T \Sigma^{-1} (y - \mu) - \tau^2\right] = 0 \label{eq:tpp_stationarity}
\end{equation}

where gradient of the ANN model $\nabla_{y_i} \hat{f}_{\text{ANN\_THR}}(y)$ and the gradient of the Mahalanobis distance constraint $\nabla_{y_i} \left[(y - \mu)^T \Sigma^{-1} (y - \mu) - \tau^2\right]$ are obtained through symbolic differentiation, while $- \mu_i^L$ and $\mu_i^U$ result from the analytical derivatives of the box constraints.$- \mu_i^L$ and $\mu_i^U$ are the multipliers of the box constraints.

Similar to the coal problem, the highly non-convex function profile of objective functions built on ANN-based models of Power and THR (data-driven approximation results in numerical instability for the convergence performance. The primal Mahalanobis distance constraint (Eq. \ref{eq:tpp_primal_maha}), its dual (Eq. \ref{eq:tpp_dual_maha}) and complementarity (Eq. \ref{eq:tpp_comp_maha}) are embedded without FB transformation. The complementarity conditions for box constraints are reformulated with $\Phi_{FB}^*$ to ensure numerical stability since constraint violations were observed without transformation. The reformulated complimentary constraints for box constraints (Eq. \ref{eq:tpp_comp_lower} - Eq. \ref{eq:tpp_comp_upper}) and inner-level constraints are given in the following:

\begin{align}
    \sqrt{(y_i)^2 + (\mu_i^L)^2 + \epsilon_l} - y_i - \mu_i^L &= 0, \quad \forall i, \label{eq:tpp_comp_lower} \\
    \sqrt{(1 - y_i)^2 + (\mu_i^U)^2 + \epsilon_u} - (1 - y_i) - \mu_i^U &= 0, \quad \forall i, \label{eq:tpp_comp_upper} \\[1em]
    (y - \mu)^T \Sigma^{-1} (y - \mu) - \tau^2 &\leq 0, \label{eq:tpp_primal_maha} \\
    \lambda_m &\geq 0, \label{eq:tpp_dual_maha} \\
    \left[(y - \mu)^T \Sigma^{-1} (y - \mu) - \tau^2\right] \cdot \lambda_m &= 0. \label{eq:tpp_comp_maha}
\end{align}

where $\epsilon_l = 10^{-3}$ and $\epsilon_u = 10^{-9}$ are small regularization parameters to smooth the complementarity constraints. The reformulated MPEC is solved using the IPOPT solver through the GAMS interface in Pyomo.

\begin{table}[ht]
\centering
\caption{Optimality, CPU Time, and Solution Quality of the Gas Power Plant Optimization under Varying Tolerance Levels}
\label{tab:gas_solutions}
\begin{tabular}{ccccc}
\toprule
Tolerance (\%) & THR (kJ/kWh) & Power (MW) & Optimality & CPU Time (s) \\
\midrule
75 & 8265 & 387 & Optimal & 0.318 \\
76 & 10265 & 401 & Infeasible & 0.493 \\
77 & 8254 & 388 & Optimal & 0.370 \\
78 & 8246 & 402 & Optimal & 0.465 \\
79 & 9011 & 388 & Infeasible & 1.161 \\
80 & 8494 & 395 & Infeasible & 0.784 \\
81 & 8230 & 386 & Optimal & 0.327 \\
82 & 10304 & 385 & Infeasible & 0.386 \\
83 & 8219 & 385 & Optimal & 0.431 \\
84 & 8208 & 399 & Optimal & 0.402 \\
85 & 8195 & 402 & Optimal & 0.383 \\
86 & 10340 & 402 & Infeasible & 0.481 \\
87 & 8175 & 381 & Optimal & 0.220 \\
88 & 8172 & 381 & Optimal & 0.466 \\
89 & 8159 & 402 & Optimal & 0.168 \\
90 & 8140 & 379 & Infeasible & 0.313 \\
91 & 8130 & 378 & Infeasible & 0.358 \\
92 & 10425 & 402 & Infeasible & 0.705 \\
93 & 8067 & 402 & Infeasible & 0.752 \\
94 & 10367 & 402 & Infeasible & 0.494 \\
95 & 8026 & 400 & Infeasible & 1.535 \\
\bottomrule
\end{tabular}
\end{table}

\section{Reformulation of bi-level Robust Optimization with KKT conditions} \label{rob_kkt}

The bi-level robust optimization problem is reformulated into a single-level with KKT-based optimality conditions written for the lower-level problem.

\begin{align}
& \max_{x, \rho, \delta, \lambda}
& & \rho \label{eq:mpec_obj} \\
& \text{subject to}
& & (x - \mu)^T \Sigma^{-1} (x - \mu) \leq \tau_{95}^2, \label{eq:mpec_upper_maha_const} \\
& & & \hat{f}_{\text{ANN}}(x + \delta) \geq TE_{\text{target}}, \label{eq:mpec_safety_floor} \\
& & & x \in [0,1]^9, \quad \rho \geq 0. \label{eq:mpec_bounds}
\end{align}

Constraint (\ref{eq:mpec_upper_maha_const}) ensures design validity by keeping the solution within the 95$^{th}$ percentile of the Mahalanobis distance of the training data. Constraint (\ref{eq:mpec_safety_floor}) enforces the safety floor, requiring the worst-case efficiency to meet the target. The KKT-conditions for the lower-level problem are written as follows:

{Stationarity Condition:}
\begin{equation}
\nabla_{\delta} \hat{f}_{\text{ANN}}(x + \delta) + \lambda \nabla_{\delta} (\delta^T \Sigma^{-1} \delta) = 0 \label{eq:mpec_stationarity}
\end{equation}

Here, $\lambda$ is the Lagrange multiplier associated with the uncertainty ellipsoid constraint $\delta^T \Sigma^{-1} \delta \leq \rho^2$. The gradients for both the neural network surrogate ($\nabla_{\delta} \hat{f}_{\text{ANN}}$) and the uncertainty set ($\nabla_{\delta} (\delta^T \Sigma^{-1} \delta)$) are computed via symbolic differentiation within the optimization framework.

{Primal Feasibility:}
\begin{align}
\delta^T \Sigma^{-1} \delta - \rho^2 &\leq 0 \label{eq:mpec_lower_ellipsoid}
\end{align}

{Complementarity Conditions:}
\begin{align}
\lambda \cdot \left(\delta^T \Sigma^{-1} \delta - \rho^2\right) &= 0 \label{eq:mpec_comp_ellipsoid}
\end{align}

{Dual Feasibility:}
\begin{align}
\lambda &\geq 0 \label{eq:mpec_dual_nonneg}
\end{align}
\end{appendices}

\end{document}